\documentclass[final,letterpaper,12pt,oneside]{class_diss}

\usepackage{graphicx} 
\usepackage{amsmath} 
\usepackage{amsxtra} 
\usepackage{amssymb} 
\usepackage{amsthm} 
\usepackage{latexsym} 
\usepackage{setspace} 
\usepackage{nomencl} 
\usepackage[margin=1in]{geometry} 
\usepackage[titles]{tocloft} 
\usepackage{arydshln}

\usepackage[super,sort&compress]{natbib}

\setcitestyle{super}

\usepackage[pdftex, plainpages=false, pdfpagelabels]{hyperref}

\hypersetup{
    linktocpage=true,
    colorlinks=true,
    citecolor=blue,
    urlcolor=blue,
    linkcolor=blue,
    citebordercolor={1 0 0},
    urlbordercolor={1 0 0},
    linkbordercolor={.7 .8 .8},
    breaklinks=true,
}

\begin{document}



\newpage

\thispagestyle{empty}

\begin{center}

\vspace{1cm}

\large Identifying bias in CNN image classification using image scrambling and transforms\\

\vspace{0.5cm}

by\\

\vspace{0.5cm}

\large Sai Teja Erukude\\

\vspace{0.3cm}

B.Tech, Jawaharlal Nehru Technological University Hyderabad, 2020\\

\vspace{0.35cm}
\rule{2in}{0.5pt}\\
\vspace{0.65cm}

{\large A THESIS}\\

\vspace{0.3cm}

\begin{singlespace}
submitted in partial fulfillment of the\\
requirements for the degree\\
\end{singlespace}

\vspace{0.3cm}
{\large MASTER OF SCIENCE}\\
\vspace{0.3cm}

\begin{singlespace}
Department of Computer Science\\
Carl R. Ice College of Engineering\\
\end{singlespace}

\vspace{0.3cm}

\begin{singlespace}
{\large KANSAS STATE UNIVERSITY}\\
Manhattan, Kansas\\
\end{singlespace}

2024\\
\vspace{0.3cm}

\end{center}

\begin{flushright}
Approved by:\\
\vspace{0.3cm}
\begin{singlespace}
Major Professor

Dr. Lior Shamir\\
\end{singlespace}
\end{flushright}



\newpage

\thispagestyle{empty}

\vspace*{0.9cm}

\begin{center}

{\bf \Huge Copyright}

\vspace{1cm}

\Large\copyright\ Sai Teja Erukude 2024.\\

\vspace{0.5cm}

\end{center}


\begin{abstract}




\pagestyle{empty}
\setlength{\baselineskip}{0.8cm}

Convolutional Neural Networks are now prevalent as the primary choice for most machine vision problems due to their superior rate of classification and the availability of user-friendly libraries. These networks effortlessly identify and select features in a non-intuitive data-driven manner, making it difficult to determine which features were most influential in the learning process. That leads to a ``black box", where users cannot know how the image data are analyzed but rely on empirical results. Therefore the decision-making process can be biased by background information that is difficult to detect. Here we discuss examples of such hidden biases and propose techniques for identifying them, methods to distinguish between contextual information and background noise, and explore whether CNNs learn from irrelevant features. One effective approach to identify dataset bias is to classify blank background parts of the images. However, in some situations a blank background in the images is not available, making it more difficult to separate the foreground information from the blank background. Such parts of the image can also be considered contextual learning, not necessarily bias. To overcome this, we propose two approaches that were tested on six different datasets, including natural, synthetic, and hybrid datasets. The first method involves dividing images into smaller, non-overlapping tiles of various sizes, which are then shuffled randomly, disrupting the features of the objects, and making classification more challenging. The second method involves the application of several image transforms, including Fourier, Wavelet transforms, and Median filter, and their combinations. These transforms help recover background noise information used by CNN to classify images. Results indicate that this method can effectively distinguish between contextual information and background noise, and alert on the presence of background noise even without the need to use background information.
\vfill
\end{abstract}


\newpage
\pagenumbering{roman}


\setcounter{page}{4}


\pdfbookmark[0]{\contentsname}{contents}


\renewcommand{\cftchapleader}{\cftdotfill{\cftdotsep}}


\renewcommand{\cftchapfont}{\mdseries}
\renewcommand{\cftchappagefont}{\mdseries}


\makenomenclature
\nomenclature{symbol}{definition}

\tableofcontents
\listoffigures
\listoftables


\newpage
\phantomsection
\addcontentsline{toc}{chapter}{List of Nomenclature}

\newpage
\vspace*{0.9cm}
\begin{center}
{\bf \Huge Nomenclature}
\end{center}

\setlength{\baselineskip}{0.8cm}



\begin{itemize}
    \item CAM - Class Activation Mappings
    \item CNN - Convolutional Neural Network
    \item DFT - Discrete Fourier Transform
    \item DWT - Discrete Wavelet Transform
    \item FC - Fully connected layer
    \item FIR - Finite Impulse Response
    \item ReLU - Rectified Linear Units
    \item ResNet - Residual Neural Network
    \item RMSprop - Root Mean Square Propagation
    \item ROI - Region Of Interest
    \item VGG16 - Visual Geometry Group 16
\end{itemize}


\newpage
\phantomsection
\addcontentsline{toc}{chapter}{Acknowledgements}

\newpage
\vspace*{0.9cm}
\begin{center}
{\bf \Huge Acknowledgments}
\end{center}

\setlength{\baselineskip}{0.8cm}

Above all, I would like to take this opportunity to be eternally grateful to my major advisor, Dr. Lior Shamir, for his outstanding mentorship and belief in my research capabilities. His knowledge and guidance have been indispensable throughout my research, giving me the confidence and shared knowledge that helped me put this thesis together. Dr. Shamir's support and constructive feedback have enriched the quality of my work and eased my path of growth considerably in both personal and professional respect.

This leads me to my committee members, Dr. Mitchell Neilsen and Dr. Torben Amtoft, to whom I give my most sincere thanks, who have so generously supported serving on my committee by devoting considerable time to this research effort. 

Lastly, I express my most profound feelings of indebtedness to my family and friends who have shared constant affection and support in this voyage. Their contribution has been fundamental in more ways than one could count. I am deeply grateful to my girlfriend, Jane Mascarenhas, whose continued love, patience, and encouragement have substantially contributed to my persistence and success.





\newpage
\pagenumbering{arabic}
\setcounter{page}{1}


\cleardoublepage

\chapter{Introduction}
\label{introduction}

Deep learning could be explained as machine learning methods that rely on deep neural networks with huge numbers of interconnected layers. These have achieved state-of-the-art performance in many other applications ranging from image classification, object detection, speech recognition, and language translation.

Unlike other forms of traditional machine learning methods, deep learning algorithms do not require explicit human involvement in feature extraction, selection, modeling, and testing. Deep learning algorithms can automatically extract and select features from raw data, like images, videos, or text. The term ``deep" in deep learning refers to many layers of algorithms, or neural networks. These are very flexible architectures that can learn directly from raw data somewhat like the human brain and, given more data, do so with improved predictive accuracy.

It is also the bedrock technology that can further reach unparalleled feasibility and accuracy in speech-to-text processing, language translation, and object recognition, to name but a few, other applications of artificial intelligence (AI). It is rightly accredited to several recent breakthroughs in the field of AI, including AlphaGo from Google DeepMind, driverless cars, voice assistants, and many more.

\begin{figure}[htb]
    \centering
    \includegraphics[height=1.5in]{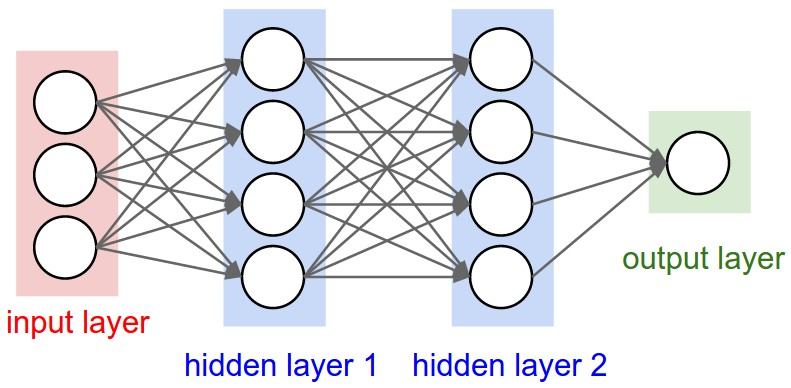}
    \caption[A deep neural network representation]
    {A deep neural network representation \cite{Abdelfattah:2017}}
    \label{fig_dnn}
\end{figure}

Deep learning uses multi-layered artificial neural networks (ANNs), which are networks composed of several ``hidden layers" of nodes between the input and output (see Figure \ref{fig_dnn}).

\section{Convolutional neural networks (CNNs)}
\label{cnns}

Convolutional neural networks, also known as ConvNets have been the most-used architecture in deep learning when it comes to image and video processing. With high intrinsic performance in feature identification and pattern identification, they are suitable for tasks such as object detection, image recognition, pattern identification, and even facial recognition. Some linear algebra practices majorly by matrix multiplication form the basis of how these CNNs search for patterns in images.

Interest in CNNs exploded with the introduction of AlexNet \cite{arXiv:1803.01164} in 2012. From that day forward, in only three short years, it went from an 8-layer AlexNet to a 152-layer ResNet \cite{Keras:2024}. The rapidity with which this occurred speaks to the massive success and popularity that both CNNs and deep learning in general are enjoying.

They work much better for high-dimensional data, including images, speech, or audio inputs, than previous methods relying on the segmentation and manual feature extraction process. They offer an image classification and object recognition approach with a scalable nature and allow for efficient data processing by reducing overfitting risks, though with the potential loss of some information in pooling layers. The main advantage of CNN models is their computational efficiency, which enables them to run on a wide range of devices.

\begin{figure}[ht]
    \centering
    \includegraphics[height=1.5in, width=6in]{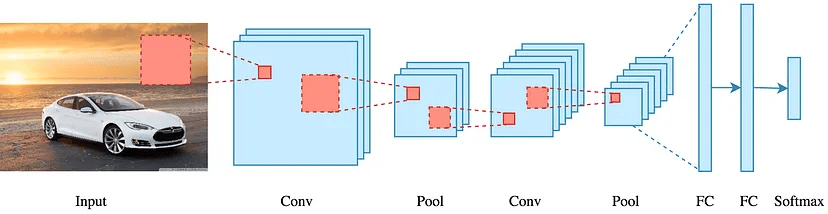}
    \caption[A convolutional neural network representation]
    {A convolutional neural network representation \cite{Dertat:2017}}
    \label{fig_cnn}
\end{figure}

Figure \ref{fig_cnn} represents the major components of CNN: convolutional layers, pooling layers, and fully connected or FC layers. A few of the CNNs are incredibly complex and involve thousands of layers, each layer enhancing the ability of the model. Early layers detect simple features of colors and edges; however, as the layers go higher, the network goes step by step to detect complex components and finally the whole object. Another strong point is that feature learning is done automatically and without intervention; this was not the case for some of the earlier models. If one trains a CNN on some pictures of cats versus dogs, it automatically learns how to select features that will make dogs distinguishable from cats.

In the hidden layers of a CNN, convolution filters are employed for feature map extraction of input data, which further undergoes processing through pooling layers and finally feeds into the output layer, marking the end of processing by the network. This is a complex process that gives fantastic results to CNNs; at the same time, it is loaded with such a level of complexity that the network seems like a "black box." What goes on within those hidden layers is very obscure, even to the creators, because of the non-intuitive rules at work.

Nowadays, these models stand at the forefront of various kinds of image-related tasks and are further expanded into many other domains such as recommender systems and natural language processing. This is the reason for the versatility and efficiency that keep them at the front line of attention when it comes to wide applicability within deep learning.

\subsection{Drawbacks of CNNs}
\label{drawbacks}

With great advantages brought about by CNN, they also bear most of the major limitations. For example, high computation degree: The computation generally involves several GPUs for execution and is a very time-consuming process, not to mention expensive. CNNs apply highly skilled experts who are knowledgeable in multidisciplinary fields to perform comprehensive testing, setting modification, and hyperparameters among many other variables.

Less discussed, however, is probably the most serious drawback: the ``black box" nature of CNNs. While CNNs are trained over data samples, updating their weights in a view to do better, the decision-making process remains opaque. The rules that define how classifications are made are convoluted and counterintuitive, and it has proven difficult to interpret the reason behind the decision of a network. The definition of what a CNN really ``learns" from the data thus becomes equally difficult to make. In other words, lack of interpretability simply means that CNNs should be used with caution as their classification processes are not certain in CNNs \cite{9154417, make3040048}. Section \ref{classification_bias} will take a deeper dive into this issue.

\section{Bias in CNN Image Classification}
\label{classification_bias}

Ideally, a CNN that can achieve high accuracy in classification on benchmark datasets should perform well in the finding of real-world objects. However, several studies proved that many popular datasets were biased and possibly not representative of a model's performance concerning real-world object recognition.\cite{TommasiPCT15}

Neural networks automatically extract and select features from an image in a non-intuitive, data-driven manner. Since the images are unsegmented and may contain irrelevant artifacts, some contributions of these images to the CNN prediction may have nothing to do with the main object of interest. It thus becomes difficult to find those concrete features that have had most of the influence on the network's learning and decision processes. Therefore, it also can be biased by the background information in such a manner that it is hard to notice.

\subsection{Benchmark datasets}
\label{benchmark_datasets}

Classification bias is present in many widely used benchmark datasets. \citet{model2015} conducted an in-depth study comparing dataset biases across various object recognition benchmarks, including COIL-20, COIL-100, NEC Animals, Caltech 101, etc. They employed a method that involved isolating a small, seemingly blank background area from each image - an area that does not contain any information about the object of interest. This technique was used to detect dataset bias in single-object recognition datasets and to compare the extent of bias across different datasets. The findings revealed that all the datasets tested still achieved classification accuracy above chance levels using these small images, even though the sub-images lacked any visually interpretable information.

The impact of bias is significantly more in object datasets collected in controlled environments, like COIL-20, COIL-100, and NEC Animals, compared to datasets sourced from natural images found online, such as ImageNet \cite{imagenet}. As these datasets achieved classification accuracy well above chance using only seemingly identical background areas indicates the presence of artifacts in the images. This suggests that similar biases may also be present in the foreground object areas, potentially enabling machine vision algorithms to classify images even without properly identifying the objects depicted.

\subsection{Medical datasets}
\label{medical_datasets}

\citet{majeed2020issues} encountered similar challenges when they investigated the applicability of CNNs for detecting COVID-19 in chest X-ray images. Their study involved using 12 established CNN architectures in transfer learning mode across three publicly available chest X-ray databases. Additionally, they introduced a custom shallow CNN model, which they trained from scratch. For their experiments, chest X-ray images were input into CNN models without any pre-processing, mirroring how other studies have used these images.

To better understand how these CNNs made their predictions, they conducted a qualitative analysis using a method called Class Activation Mappings (CAM) \cite{winastwan2024cam}. CAMs allow researchers to generate heatmaps of the most discriminating regions and visualize which parts of the input image were most influential in the network's decision-making process. By mapping the activations from the CNN back to the original image, one can highlight class-specific regions that contributed most significantly to the model's classification, providing insights into which features were most relevant for detecting COVID-19.

\begin{figure}[ht]
    \centering
    \includegraphics[width=6.5in]{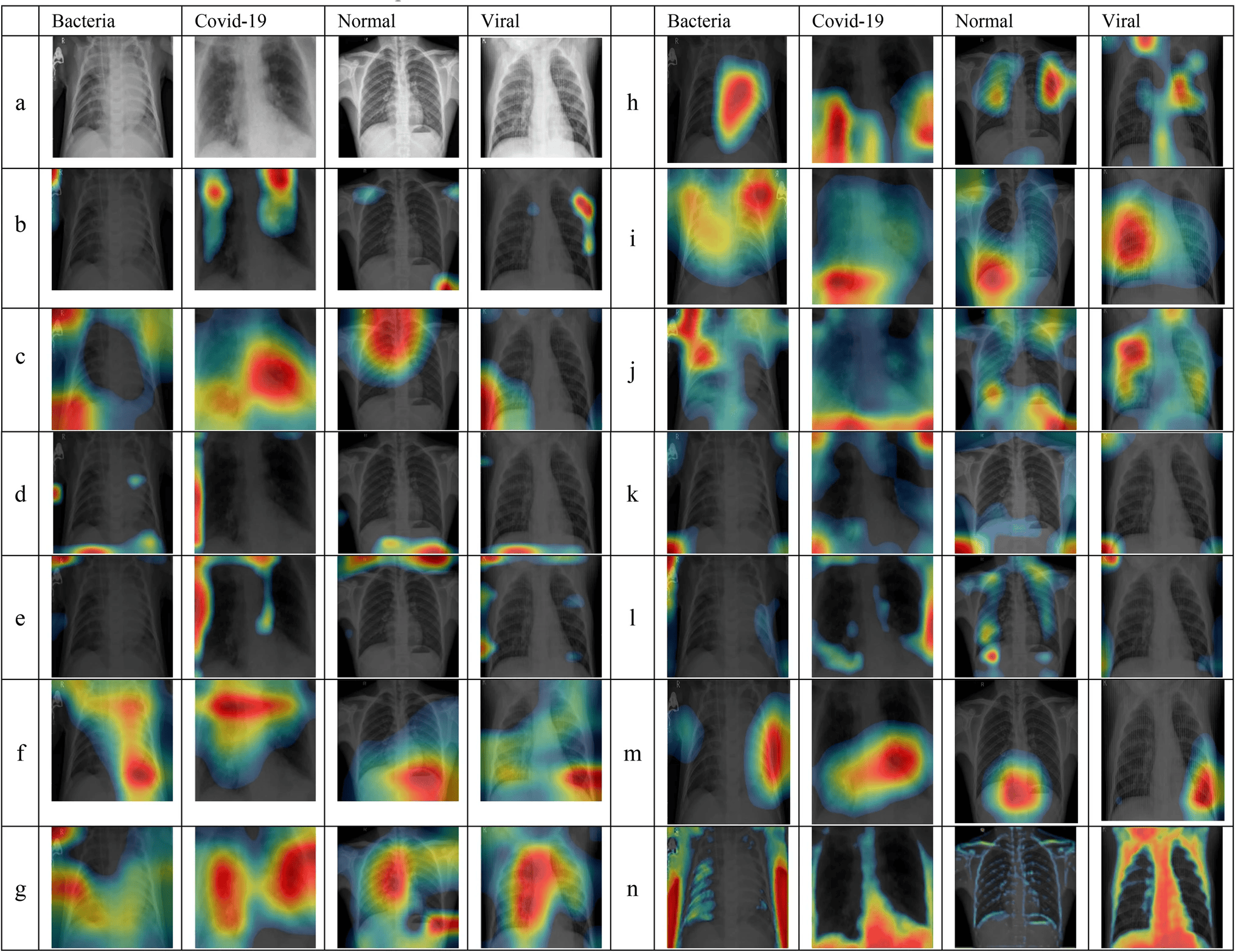}
    \caption[X-rays images classified correctly by CNNs]
    {X-rays images classified correctly by CNNs. a Original X-ray, b AlexNet, c GoogleNet, d VGG16, e VGG19, f ResNet18, g ResNet50, h ResNet101, i Inception V3, j InceptionResNet, k DenseNet201, l SqueezeNet, m Xception, and n CNN-X \cite{majeed2020issues}}
    \label{fig_13models_cam}
\end{figure}

Figure \ref{fig_13models_cam} illustrates the regions of images that CNN models focused on when making predictions. It is evident that, in only a few cases, the models concentrated on the frontal chest region (i.e., the lung area), which is where we typically look for signs of COVID-19 and other infections. Instead, these models often focused on areas outside the chest's frontal view, as seen in the first column of row (b) and the third and fourth columns of row (e) in Figure 5. Additionally, there are noticeable overlaps between the CAM hot spots and textual elements in the images, particularly in the first column of row (b), and the first, third, and fourth columns of row (e).

This is a notable example where almost all of the CNN architectures used areas outside the region of interest (ROI) to make their final classification predictions. This observation leads us to conclude that directly using X-ray images without pre-processing such as segmenting the ROI and removing background noise, can lead to biased and misleading classification results.

In summary, CNN predictions should be considered cautiously, even if they have high accuracy until the region(s) of the input image used by CNNs that lead to its prediction are visually inspected and approved. It is also demonstrated that CNNs are ‘cheating’ by using artifacts like text or medical device marks and background noise present in the images to build their prediction that has nothing to do with the object of interest.

It is now evident that nearly every dataset contains some level of bias, and this should be taken into account before relying solely on high-accuracy figures. This thesis proposes several techniques for detecting classification bias, even in cases where it is not feasible to isolate segments of the image background.
\cleardoublepage

\chapter{Data and CNN architecture}
\label{data_cnn_architecture}

\begin{figure}[ht]
    \centering
    \includegraphics[width=6.3in]{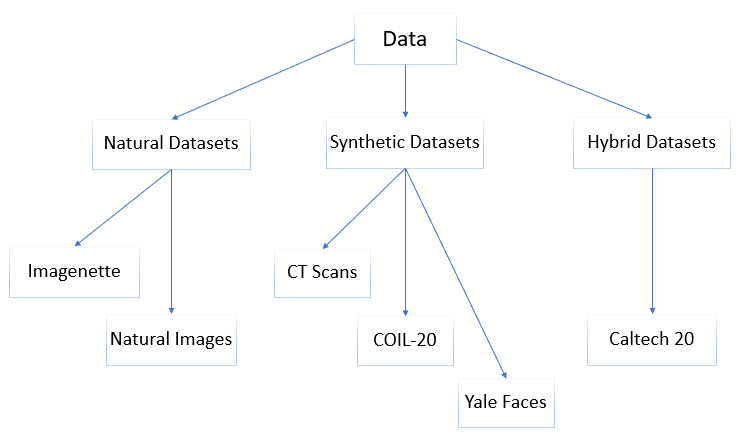}
    \caption[Categorization of data]
    {Categorization of data}
    \label{fig_data_categories}
\end{figure}

This research was carried out across six distinct datasets to ensure that the findings were valid and robust across different contexts. As shown in Figure \ref{fig_data_categories}, datasets used in this study can be further categorized into three types: natural datasets \ref{natural_datasets}, non-natural or synthetic datasets \ref{non-natural_datasets} and mixed or hybrid datasets \ref{mixed_datasets}. Successive sections describe in detail the source of the datasets and the sizes of the training, testing, and validation splits.

\section{Natural datasets}
\label{natural_datasets}

Natural datasets involve data in which, in general, information is taken from the real world, representing some natural phenomenon or environment. They contain images captured in real life, like photographs of people, animals, objects, and scenes. They have high variability in lighting conditions, angles, backgrounds, and object appearances that help in making models robust for real-world conditions. These often contain a variety of patterns and complex noise and variations that make the training of the model more difficult, but create better applicability for real-world tasks. For example, two such naturally collected datasets used in this work are Imagenette \ref{imagenette} and Natural images \ref{natural_images}.

\subsection{Imagenette}
\label{imagenette}

\begin{figure}[ht]
    \centering
    \includegraphics[width=5.8in]{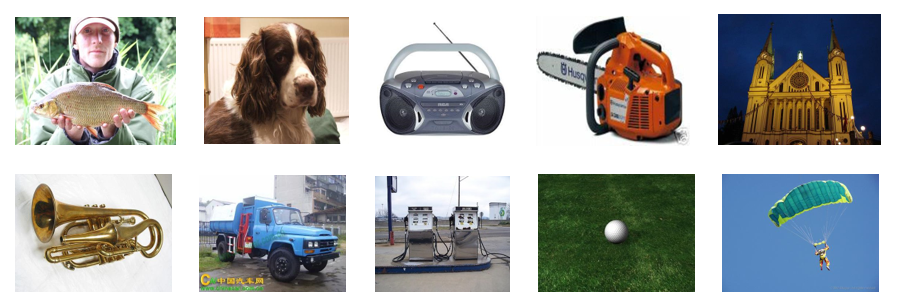}
    \caption[{\it Imagenette} dataset with 10 classes]
    {{\it Imagenette} dataset with 10 classes}
    \label{fig_imagenette}
\end{figure}

{\it Imagenette} \cite{yongwo:2024} is a smaller subset of the {\it ImageNet} \cite{imagenet} dataset, and it contains 10 different classes. It is great to quickly experiment and fine-tune techniques before applying them to the full-scale {\it ImageNet} dataset. Table \ref{table_imagenette} illustrates the train, test, and val split and the figure \ref{fig_imagenette} depicts the 10 different classes present in the dataset.

\begin{table}[ht]
    \setlength{\tabcolsep}{16pt} 
    \renewcommand{\arraystretch}{1.6} 
    \begin{center}
    \begin{tabular}[c]{|c|c|c|c|}
        \hline
        Train & Test & Validation & Total Images\\
        \hline
        9,469 & 1,904 & 2,021 & 13,394\\
        \hline
    \end{tabular}
    \caption{Split of Imagenette dataset}
    \label{table_imagenette}
   \end{center}
\end{table}

\subsection{Natural images}
\label{natural_images}

The {\it Natural images} dataset consists of images coming from 8 different classes collected from several sources \cite{roy2018effects}. The classes include airplane, car, cat, dog, flower, fruit, motorbike and person. The class ``fruit" will be discarded for the purposes of this research because the background was changed to white, which could introduce biases in the classification process.

\begin{figure}[ht]
    \centering
    \includegraphics[width=5.8in]{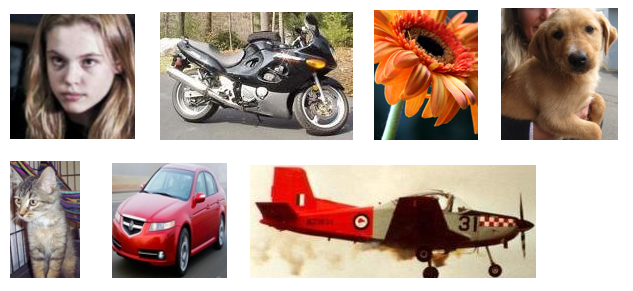}
    \caption[{\it Natural images} dataset]
    {{\it Natural images} dataset}
    \label{fig_natural_images}
\end{figure}

Table \ref{table_natural_images} shows the distribution of the dataset across training, testing and validation sets while Figure \ref{fig_natural_images} shows all 7 subjects.

\begin{table}[ht]
    \setlength{\tabcolsep}{16pt} 
    \renewcommand{\arraystretch}{1.6} 
    \begin{center}
    \begin{tabular}[c]{|c|c|c|c|}
        \hline
        Train & Test & Validation & Total Images\\
        \hline
        3,367 & 709 & 711 & 4,787\\
        \hline
    \end{tabular}
    \caption{Split of {\it Natural images} dataset}
    \label{table_natural_images}
   \end{center}
\end{table}

\section{Non-natural or synthetic datasets}
\label{non-natural_datasets}

Non-natural datasets, on the other hand, consist of data that are either synthetically generated or created under controlled conditions. Since these datasets are typically generated in controlled environments or through algorithms, they exhibit less variability compared to natural datasets. They are frequently used to streamline model training or to emphasize particular characteristics of the data. Designed to address specific tasks or problems, such as recognizing patterns in simplified scenarios, these datasets tend to have simpler features and reduced variability, making them useful for targeted experiments or testing specific hypotheses. 

Examples of synthetic datasets include images generated through computer graphics or simulation tools, such as those from 3D rendering engines, virtual environments, or simulations of medical imaging and robotic vision. In this research, we used three such non-natural datasets, which will be described in the following sections. Studies have demonstrated that all three datasets contain some form of bias or background noise, which impacts the learning process of CNNs.

\subsection{CT scans}
\label{ct_scans}

This dataset consists of CT scan images of lungs that are categorized into four classes: COVID-19, normal, bacterial pneumonia, and viral pneumonia. Note that these images have many types of artifacts including text, medical device traces, and background noise which will significantly influence the CNN model's learning and decision-making process \cite{majeed2020issues}. Additionally, it was demonstrated that CNNs were using regions/features in the input image that are outside the region of interest (ROI) and have no relation with COVID-19.

\begin{figure}[!ht]
    \centering
    \includegraphics[width=4.1in]{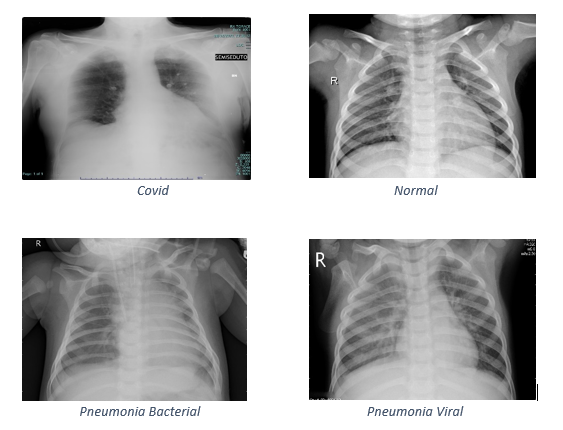}
    \caption[CT scans dataset with 4 classes]
    {CT scans dataset with 4 classes}
    \label{fig_ct_scans}
\end{figure}

Table \ref{table_ct_scan} shows the distribution of these images across training, testing, and validation sets, while Figure \ref{fig_ct_scans} displays examples of the four distinct classes within the dataset.

\begin{table}[ht]
    \setlength{\tabcolsep}{16pt} 
    \renewcommand{\arraystretch}{1.5} 
    \begin{center}
    \begin{tabular}[c]{|c|c|c|c|}
        \hline
        Train & Test & Validation & Total Images\\
        \hline
        1,188 & 224 & 226 & 1,638\\
        \hline
    \end{tabular}
    \caption{Split of CT scan dataset}
    \label{table_ct_scan}
   \end{center}
\end{table}

\newpage
\subsection{Coil-20}
\label{coil-20}

{\it Coil-20}, short for Columbia object image library \cite{nene1996columbia} is a repository of gray-scale pictures that contains such twenty objects. All the objects could be precisely put on a turntable. The turntable was turned at 360 degrees until a change of object was appropriately visible from a static camera. Images of the objects were captured in 5-degree angular orientations. \citet{model2015} demonstrated in their study that {\it COIL-20}, being generated in a controlled environment, contains biases that significantly affect the model's classification accuracy. They achieved exceptionally high classification accuracy even when the object was absent from the testing image, which means the model has been memorizing patterns from bias or background noise.

Table \ref{table_coil-20} shows the distribution of the data across training, testing, and validation sets, while Figure \ref{fig_ct_scans} displays examples of the twenty different classes within the dataset.

\begin{table}[ht]
    \setlength{\tabcolsep}{16pt} 
    \renewcommand{\arraystretch}{1.6} 
    \begin{center}
    \begin{tabular}[c]{|c|c|c|c|}
        \hline
        Train & Test & Validation & Total Images\\
        \hline
        1,160 & 140 & 140 & 1,440\\
        \hline
    \end{tabular}
    \caption{Split of Coil-20 dataset}
    \label{table_coil-20}
   \end{center}
\end{table}

\begin{figure}[ht]
    \centering
    \includegraphics[width=4.3in]{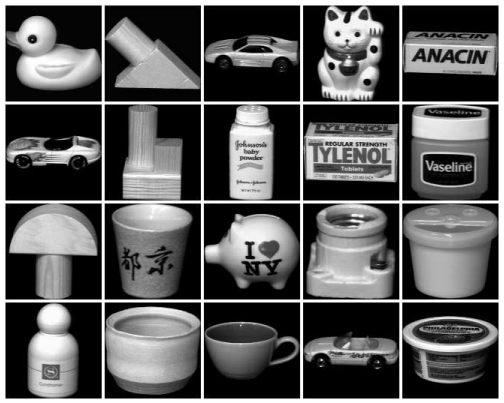}
    \caption[Coil-20 dataset with 20 classes]
    {Coil-20 dataset with 20 classes}
    \label{fig_coil-20}
\end{figure}

\subsection{Yale faces}
\label{yale_faces}

This dataset is originally from the {\it Yale Face} Database. \cite{georghiades2001few, olga:2018} There are 165 GIFs of 15 subjects, containing the following facial expressions: happy, sad, with and without glasses, sleepy, normal, wink, surprised, and lighting from different directions. \citet{sanchari2021} assessed benchmark datasets, including {\it Yale Faces}, to evaluate CNN performance. Their findings reveal a consistent bias within the dataset, with the CNN achieving notably higher classification accuracy when only a part of the image was used, highlighting the significant extent of this bias.

\begin{figure}[ht]
    \centering
    \includegraphics[width=6.5in]{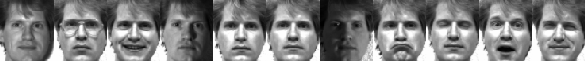}
    \caption[Yale faces dataset]
    {Yale faces dataset}
    \label{fig_yale_faces}
\end{figure}

Figure \ref{fig_yale_faces} displays all the images of a subject. The GIFs were then converted to JPEG format and augmented with horizontal and vertical flips. Table \ref{table_yale_faces} provides the distribution of the augmented dataset across training, testing, and validation sets.

\begin{table}[ht]
    \setlength{\tabcolsep}{16pt} 
    \renewcommand{\arraystretch}{1.6} 
    \begin{center}
    \begin{tabular}[c]{|c|c|c|c|}
        \hline
        Train & Test & Validation & Total Images\\
        \hline
        315 & 90 & 90 & 495\\
        \hline
    \end{tabular}
    \caption{Split of augmented Yale faces dataset}
    \label{table_yale_faces}
   \end{center}
\end{table}

\section{Mixed or Hybrid datasets}
\label{mixed_datasets}

Mixed or hybrid datasets are collections of images that combine both real-world (natural) images and synthetically generated images. Such datasets are designed for leveraging both data types to improve the performance and generalization of models in tasks such as image classification. Hybrid datasets offer a wider variety of scenarios and conditions, which helps models to generalize more effectively. Additionally, the ability to generate synthetic images in large quantities helps mitigate the problem of having limited real-world data.

\subsection{Caltech 20}
\label{caltech_256}

The {\it Caltech 20} database is a subset of the larger {\it Caltech 256} dataset, which was sourced from Caltech Data and comprises 25,607 images across 256 object categories. For the purpose of this study, only the first 20 object categories were used to simplify and expedite the process. Interestingly, upon closer inspection, we found that the first 20 classes feature a combination of real-world images and either graphically generated or have had their backgrounds altered to a plain white/black backdrop. Therefore, {\it Caltech 20} is categorized as a mixed dataset throughout this study.

\begin{figure}[ht]
    \centering
    \includegraphics[width=5in]{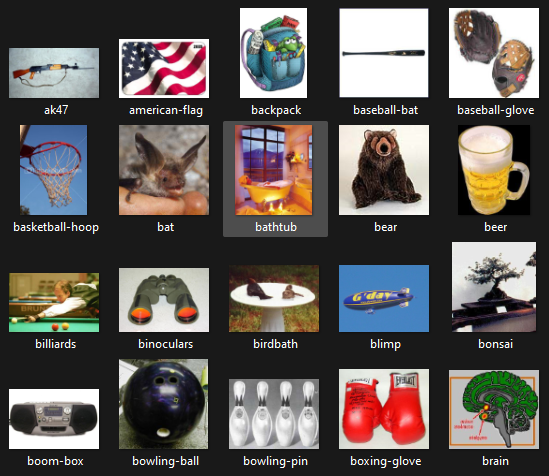}
    \caption[Caltech 20 dataset]
    {Caltech 20 dataset}
    \label{fig_caltech_20}
\end{figure}

Table \ref{table_caltech_20} provides the distribution of the augmented dataset across training, testing, and validation sets, and Figure \ref{fig_caltech_20} displays all 20 subjects.

\begin{table}[ht]
    \setlength{\tabcolsep}{16pt} 
    \renewcommand{\arraystretch}{1.6} 
    \begin{center}
    \begin{tabular}[c]{|c|c|c|c|}
        \hline
        Train & Test & Validation & Total Images\\
        \hline
        1,812 & 368 & 360 & 2,540\\
        \hline
    \end{tabular}
    \caption{Split of Caltech 20 dataset}
    \label{table_caltech_20}
   \end{center}
\end{table}

\section{CNN architecture}
\label{cnn_architecture}

This study utilized the VGG16 convolutional neural network, which stands for ``Visual Geometry Group 16." Introduced by \citet{arXiv:1409.1556} from the University of Oxford in their 2014 paper ``Very Deep Convolutional Networks for Large-Scale Image Recognition," this model achieved an impressive accuracy of 92.77\% on the ImageNet dataset that contained 14 million images falling under 1000 classes.

\begin{figure}[ht]
    \centering
    \includegraphics[width=6.5in]{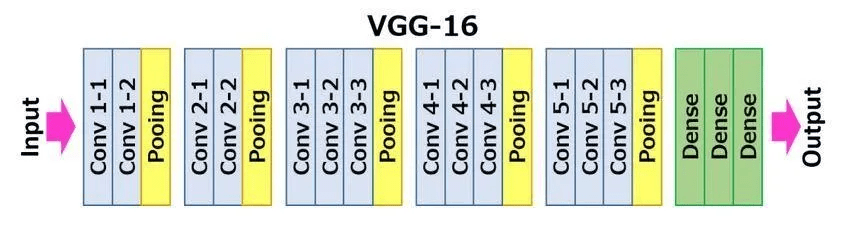}
    \caption[VGG16 architecture map]
    {VGG16 architecture map}
    \label{fig_vgg16_outline}
\end{figure}

The ``16" in VGG16 refers to the network's 16 weight layers, which are the layers that have learnable parameters. Specifically, VGG16 comprises 13 convolutional layers, five max-pooling layers, and three fully connected (dense) layers, adding up to a total of 21 layers, but only 16 of these layers have weights (see Figure \ref{fig_vgg16_outline}). The network takes as input images of size 224$\times$224 with three RGB channels. 

\begin{figure}[ht]
    \centering
    \includegraphics[width=6.5in]{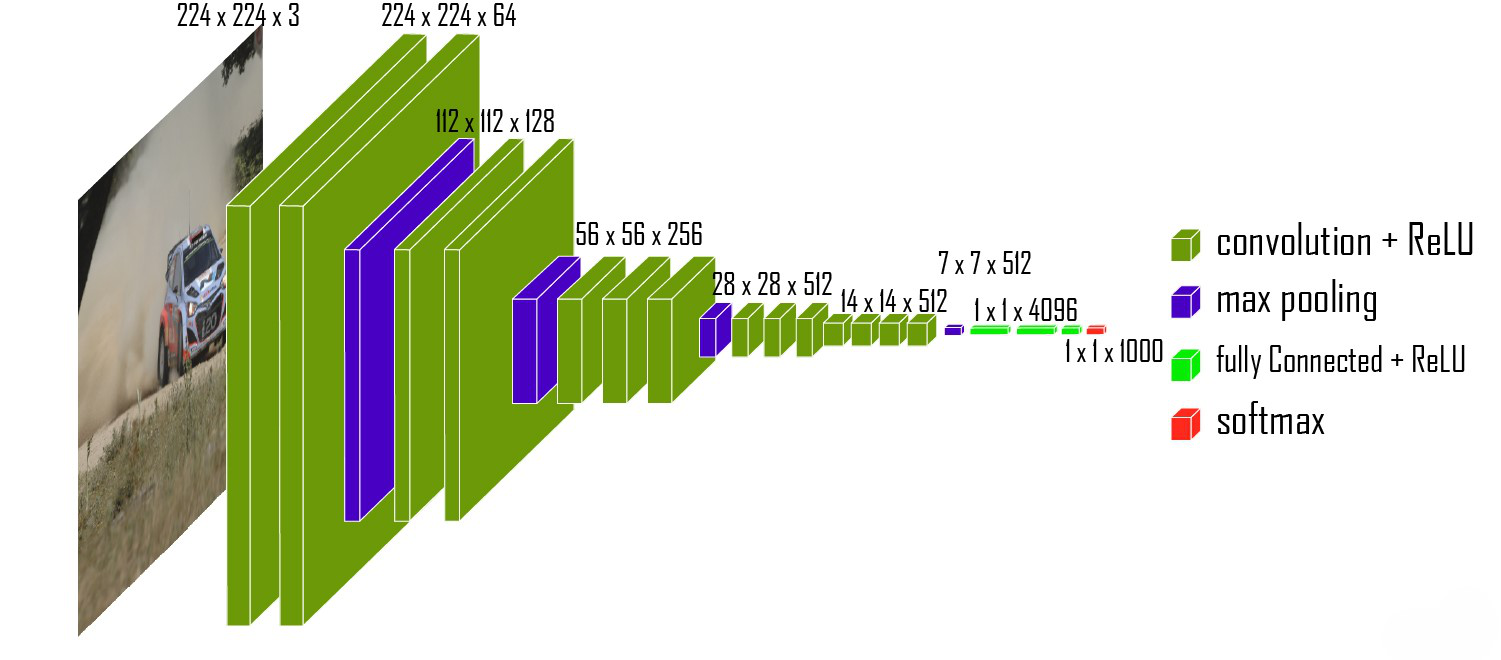}
    \caption[VGG16 architecture]
    {VGG16 architecture}
    \label{fig_vgg16_architecture}
\end{figure}

An impressive fact about the VGG16 is that its architecture is as simple as it gets: a bunch of convolutional layers with 3 × 3 filters and stride 1, while the padding is always the same, coupled with max-pooling layers with 2 × 2 filters and stride 2. This consistent reliance on small filters and pooling layers throughout the network is a key characteristic of VGG16. As illustrated in Figure \ref{fig_vgg16_architecture}, Conv-1 has 64 filters, Conv-2 has 128, Conv-3 has 256, and both Conv-4 and Conv-5 have 512 filters each. After these convolutional layers, the network includes three fully connected layers: the first two have 4096 channels each, and the third layer, which is designed for classification, has several channels matching the number of classes in the dataset. 

The network primarily uses ReLU, short for rectified linear unit as the activation function, except in the final layer where a softmax function is used to produce probabilities for each class. An activation function is a mathematical function that introduces non-linearity allowing the network to learn complicated patterns and mappings. Without it, the network would just be a linear function which is a polynomial of degree one \cite{sharma2017activation}.\\ 

The optimizer utilized here is RMSprop, which stands for Root Mean Square Propagation. The learning rate here was taken as 0.00001. Optimizers are functions that change neural network attributes: weights and learning rates to decrease the total loss and increase accuracy. A comprehensive comparison study of the most used optimizers in neural networks has been conducted by \citet{desai2020comparative}.
\cleardoublepage

\chapter{Identifying bias using background segments and image scrambling}
\label{identifying_bias_image_scrambling}

\section{Introduction}
\label{indentifying_bias_introduction}

\subsubsection{What is bias?}
In the context of CNN image classification, bias is a case of systematic inaccuracies in model predictions due to certain patterns or irregularities within either data or the learning process of a model. This kind of bias would finally affect the way CNN learns to recognize classes and finally result in prejudiced or incorrect classifications.

\subsubsection{How are biases caused?}
Biases can emerge as early as the data acquisition stage and can be influenced by various factors, which are elaborated on in the following discussion:

\begin{enumerate}
    \item \textbf{Capture Bias:} Despite the best efforts of their creators, the datasets can have a strong built-in bias. For example, datasets captured in a controlled environment, where all images are taken during a single session, may exhibit inherent biases. These biases might be due to consistent artifacts or characteristics from the session rather than from the objects themselves. A prominent type of bias is that objects are almost always positioned in the center of images. If you search for "mug" on Google Image Search, you will notice another kind of capture bias: nearly all the mugs feature a handle that is oriented to the right. Subtle variations such as changes in lighting conditions or camera temperature can create artifacts that are hard to notice visually but will strongly influence CNN's learning process.

    \item \textbf{Selection Bias:} This type has been particularly prevalent in World Wide Web-downloaded datasets, where photo acquisition is uncontrolled. Indeed, using only an internet search often fails to provide a representative sample because the keyword-based search returns only certain types of images. \cite{torralba2011unbiased} In order to reduce selection bias, it helps to collect data by combining sources: for example, taking images using multiple search engines from different countries. \cite{imagenet}
        
    \item \textbf{Irrelevant Features:} CNNs may learn from the background or artifacts in the images rather than focusing on the subject of interest itself. For example, if all images of a dog are presented against a specific background, then the model might learn to associate the background with the dog, rather than recognizing the dog itself.

    \item \textbf{Imbalanced Datasets:} This is when some classes are dominating while others are not. CNN could be biased towards those classes that occur more frequently. This leads to its poor performance in less common classes, and it could not generalize well for all scenarios.
\end{enumerate}

\subsubsection{How do biases impact the classification?}
\begin{enumerate}
    \item \textbf{Impacted Decision-Making:} Dataset biases have a substantial impact on the neural network's learning and decision-making. As a result, the model may end up focusing on artifacts, background noise, or biases introduced during the data acquisition process, causing the main object of interest to become secondary or ignored. As a result, the model may base its classifications on these biases rather than the primary object.

    \item \textbf{Limited Generalization:} Models might excel with training data but struggle with real-world data or diverse test cases because they depend on biased features.

    \item \textbf{Unreliable Classification:} Even with high classification accuracy, predictions can be unreliable if the CNN focuses on areas that are not the primary object of interest. Bias can also lead to unequal performance across different classes, resulting in unfair predictions.
\end{enumerate}
Therefore, it becomes crucial to identify any classification or dataset biases and be cautious about interpreting the classification results.

\subsubsection{How to identify bias?}
There are several methods to identify if your dataset has a consistent bias that is influencing CNN's classification:

\begin{enumerate}
    \item \textbf{Cropping Background Segments:} This method involves cropping a small, typically blank section of the background that lacks visual information about the object of interest and then classifying the CNN based on this cropped dataset. If the CNN can accurately predict the class of the image from just this small background portion, it indicates the presence of consistent bias in the dataset, explained in greater detail in section \ref{cropping_background_segments}.

    \item \textbf{Image Scrambling:} When the background is inseparable from the foreground and cropping is not feasible, the scrambling method is highly effective. This technique involves dividing the image into small tiles, which are then randomly shuffled, disrupting the image's patterns. Details of this method are discussed in section \ref{image_scrambling}.

    \item \textbf{Image Transforms:} Applying image transformations like Fourier, Wavelet, and Median filters during the pre-processing stage can smooth out datasets and enhance model robustness. The Wavelet transform, in particular, handles natural and synthetic datasets differently, offering a method for detecting biases. This process is explained in detail in chapter \ref{identifying_bias_image_transforms}.

    \item \textbf{Class Activation Mappings (CAM) \cite{winastwan2024cam}:} Generates heatmaps to show which parts of the input image were most influential in the network's decision-making. By mapping CNN activations back to the original image, it highlights the regions that most significantly impacted the model’s classification.

    \item \textbf{Performance Evaluation:} Analyze the confusion matrix to identify patterns of misclassification and whether certain classes contain consistent bias. Additionally, test the CNN on different subsets of data or under varied conditions to see if performance varies significantly.
\end{enumerate}




\section{Methods}
\label{identifying_methods}

In this chapter, we aim to identify any persistent biases in datasets through two efficient techniques, cropping background segments \ref{cropping_background_segments} and image scrambling \ref{image_scrambling}. These methods will help us investigate whether CNN is learning primarily from the object of interest or from irrelevant background features. This is crucial for assessing the reliability of the classification accuracy reported by the model. A high classification accuracy might be misleading if the CNN is relying on the bias or background patterns or other non-essential features rather than the core characteristics of the object. 

We encourage everyone to follow these methods outlined in this thesis to analyze your dataset for any such biases or background noise or misleading features that CNN might be picking up.

\subsection{Cropping background segments}
\label{cropping_background_segments}

\begin{figure}[ht]
    \centering
    \includegraphics{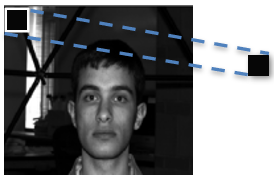}
    \caption[20$\times$20 cropped background segment]
    {20$\times$20 cropped background segment}
    \label{fig_cropping_bg}
\end{figure}

\citet{sanchari2021} have proposed a method that entails extracting a small segment from the background of an image, specifically a 20$\times$20 pixel area from the top-left corner that does not include the object of interest, as shown in Figure \ref{fig_cropping_bg}. These background segments are usually blank and devoid of any significant visual details. By compiling a dataset from these cropped, blank background portions and training a CNN to classify them, we can determine if the model is subconsciously learning from irrelevant background patterns rather than focusing solely on the object itself. Ideally, background features should have no influence on the classification process.

\begin{figure}[ht]
    \centering
    \includegraphics{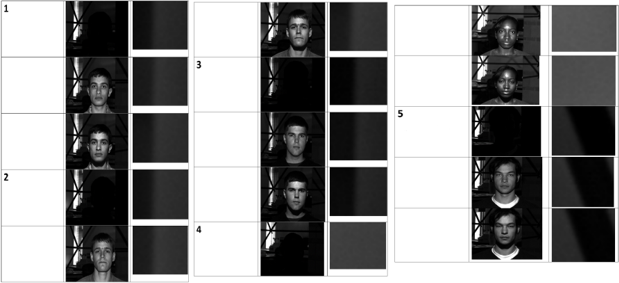}
    \caption[20$\times$20 cropped yale faces dataset]
    {20$\times$20 cropped yale faces \cite{georghiades2001few} dataset}
    \label{fig_cropped}
\end{figure}

Figure \ref{fig_cropped} shows the original Yale Faces \cite{griffin_holub_perona_2022} dataset alongside the newly generated 20$\times$20 cropped dataset. A CNN has been trained and tested on both the original and the cropped datasets, and the intriguing results are discussed in section \ref{cropped_background_segments_results}.

\begin{figure}[ht]
    \centering
    \includegraphics{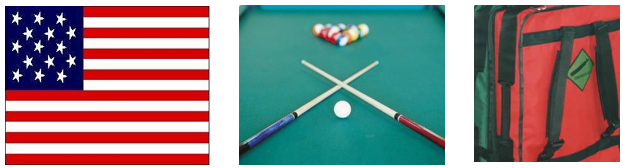}
    \caption[Images with no background]
    {Images with no background}
    \label{fig_images_without_bg}
\end{figure}

The drawback of this approach is twofold: firstly, a blank background may not always be readily available, making it hard to separate foreground information. It is not feasible when the entire image is a subject of interest and cropping a part of it is not an option. Figure \ref{fig_images_without_bg} shows examples of images where this method would be ineffective due to the lack of a background.

Secondly, using a blank background might lead to confusion with contextual learning. These limitations can restrict the flexibility and feasibility of implementing the solution. However, these problems can be addressed through the use of the image scrambling technique described in section \ref{image_scrambling}.

\subsection{Image scrambling}
\label{image_scrambling}

Image scrambling entails the arbitrary rearrangement of pixels to render the image unrecognizable and disrupt the correlation among the adjacent pixels. It is important to note that while the pixel values themselves remain unchanged during this process, their positions are altered. Image pixels are highly correlated with neighboring pixels and a good scrambling algorithm reduces this correlation near to zero. \citet{Mondal:2019}, \citet{modak2015comprehensive} discusses a range of scrambling techniques, such as matrix transformation, bit scrambling, pixel swapping, row and column swapping, Rubik's cubic algorithm, and Arnold's transformation.

\begin{figure}[ht]
    \includegraphics[width=6.5in]{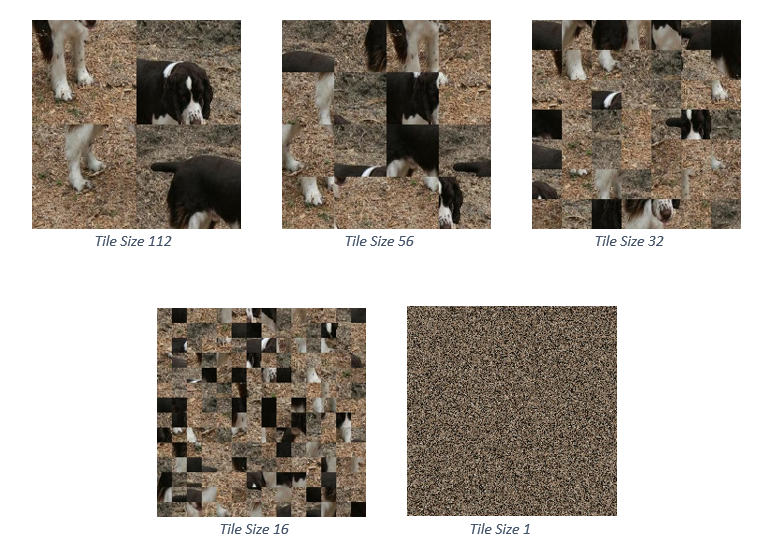}
    \caption[Tile scrambling technique outputs]
    {Image scrambling technique outputs}
    \label{fig_scrambled}
\end{figure}

In this research, a very straightforward technique called ``Tile scrambling" was used. This method involves dividing an image into smaller, non-overlapping tiles and then shuffling these tiles in a random order. In order to assess the effectiveness of scrambling, many tile sizes were tested as shown in Figure \ref{fig_scrambled}. This approach allows for the assessment of how well the model can handle changes in image structure and disrupt patterns formed by neighboring pixels.

When an image lacks a clear background or when cropping out a section of the background is not feasible, image scrambling is highly recommended as a technique to assess dataset bias and evaluate CNN performance. A CNN has been trained and tested using this scrambled dataset, and the results of this process are discussed in section \ref{image_scrambling_results}.

\newpage
\section{Results}
\label{results}

\subsection{Classification results}
\label{classification_results}

VGG16 models were trained and evaluated on all six datasets described in section \ref{data_cnn_architecture} without any pre-processing. Table \ref{table_raw_results} presents the classification results for the models tested on the raw datasets. The results show that our model accurately predicts the classes and achieves high classification accuracy across all six datasets. Notably, the model performed exceptionally well with the Coil-20 dataset, achieving an accuracy of \(\sim\)100\%, which raises concerns about the model possibly memorizing dataset-specific biases due to Coil-20 being synthetically generated. On the other hand, the Caltech 20, which features a combination of natural and non-natural images, showed a lower accuracy of around \(\sim\)36.6\%, reflecting the increased variability and complexity of this dataset compared to others.

\begin{table}[ht]
    \renewcommand{\arraystretch}{1.6} 
    \begin{center}
    \begin{tabular}[c]{|c|c|c|c|c|c|}
        \hline
        Dataset & Classification accuracy & Random accuracy\\
        \hline
        Imagenette  & 59\%      & 10\%      \\
        Natural images & 85\%    & 14.28\%  \\
        \hdashline
        CT scans    & 70\%      & 25\%      \\
        Coil-20     & 100\%     & 5\%       \\
        Yale Faces  & 70\%      & 6.66\%    \\
        \hdashline
        Caltech 20  & 36.6\%    & 5\%       \\
        \hline
    \end{tabular}
    \caption{Classification accuracies of raw datasets}
    \label{table_raw_results}
   \end{center}
\end{table}

In the subsequent sections, we will investigate whether these classification accuracies exhibit any biases or if background patterns are influencing the model’s learning process. This analysis will help us assess the reliability of these accuracies and determine the trustworthiness of our classification model.

\subsection{Cropped background segments results}
\label{cropped_background_segments_results}

\begin{figure}[ht]
    \centering
    \includegraphics{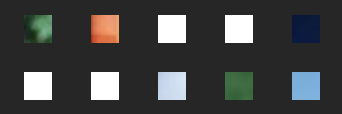}
    \caption[Corresponding 20$\times$20 cropped Imagenette dataset]
    {Corresponding 20$\times$20  cropped Imagenette dataset}
    \label{fig_cropped_imagenette}
\end{figure}

The cropping background segments method discussed in Section \ref{cropping_background_segments} has been applied to all six datasets. The output of applying this method to the Imagenette dataset (Figure \ref{fig_imagenette}), is depicted in Figure \ref{fig_cropped_imagenette}. The cropped images lack any features related to the object, are highly similar, and are indistinguishable from the naked eye.

A VGG16 model detailed in Section \ref{cnn_architecture} was trained and tested on all the datasets, including both the original and cropped versions. Table \ref{table_cropped_results} presents the classification accuracies when trained on the cropped datasets, compared to those trained on the original datasets.

\begin{table}[ht]
    \renewcommand{\arraystretch}{1.6} 
    \begin{center}
    \begin{tabular}[c]{|c|c|c|c|c|c|}
        \hline
        Dataset & Full image accuracy & Cropped accuracy & Random accuracy\\
        \hline
        Imagenette  & 59\%   & 17\%   & 10\%      \\
        Natural images  & 85\% & 27.6\% & 14.28\%       \\
        \hdashline
        CT scans    & 70\%   & 44\%   & 25\%      \\
        Coil-20     & 100\%  & 27.8\% & 5\%       \\
        Yale Faces  & 70\%   & 17.7\% & 6.66\%    \\
        \hdashline
        Caltech 20  & 36.6\% & 12.7\% & 5\%       \\
        \hline
    \end{tabular}
    \caption{Classification accuracies of cropped datasets}
    \label{table_cropped_results}
   \end{center}
\end{table}

\begin{figure}[ht]
    \includegraphics[width=6.5in]{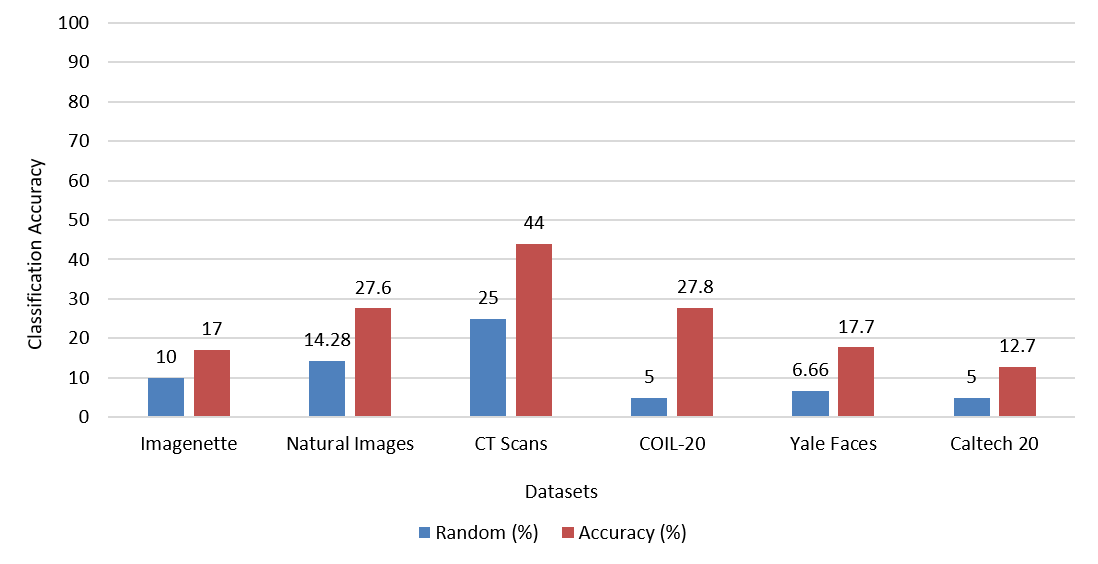}
    \caption[Classification accuracies of cropped segments vs random chance]
    {Classification accuracies of cropped segments vs random chance}
    \label{fig_cropped_graph}
\end{figure}

It is evident from the results that models trained and tested on 20$\times$20 pixel cropped datasets achieved much higher classification accuracy than random guessing in all cases. The accuracies for cropped images are consistently at least twice as high as the random accuracies across all datasets (see Figure \ref{fig_cropped_graph}). This suggests that the CNN can accurately predict the object's class by examining the top left 20$\times$20 pixel section of the image, which lacks the object's features and is quite uniform across different images.

The results reveal that the Coil-20 dataset, which was synthetically produced in a controlled environment show significantly higher classification accuracy compared to datasets gathered from diverse sources where images were not specifically intended for object recognition tasks. The random guessing accuracy of 5\% was improved by nearly 550\%, reaching  \(\sim\)27.8\% when using just a 20$\times$20 pixel region from each original image. Similarly, the Yale Faces dataset with an initial random guessing accuracy of 6.66\%, saw an increase of about 260\%, rising to around \(\sim\)17.7\%. These findings suggest that image features that are irrelevant for object recognition but help distinguish between image classes are more prevalent in non-natural datasets, while such features are less dominant in datasets collected from a variety of sources.

A possible explanation for such high cropped accuracy is that the background contains a strong signal, which significantly influences and creates a consistent bias in the network's learning process. This bias, though not directly related to the object’s features, can distinguish between image classes, resulting in better-than-expected performance.

\subsection{Image scrambling results}
\label{image_scrambling_results}

The image scrambling technique outlined in Section \ref{image_scrambling} was applied to all six datasets. Three tile sizes were used in this experiment: 1, 16, and 32. With tile sizes of 1 and 16, the scrambled images are virtually unrecognizable to the human eye and difficult to classify. Tile size 32 offers some level of guess-ability, though it is not consistently reliable. Figure \ref{fig_scrambled} provides a clear comparison of the effects of each tile size.

\begin{figure}[ht]
    \centering
    \includegraphics[width=6in]{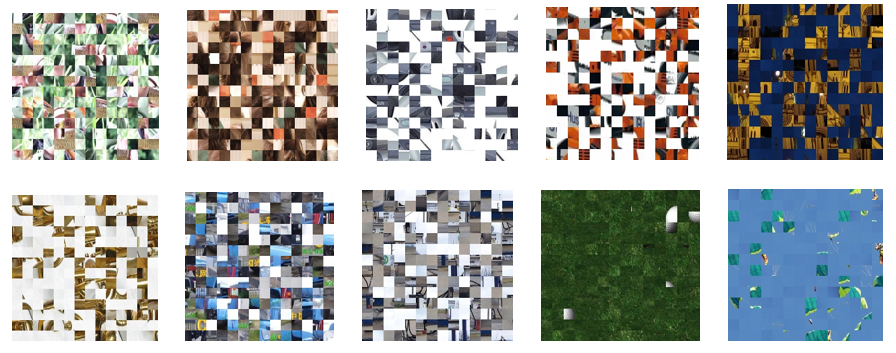}
    \caption[Corresponding scrambled Imagenette dataset]
    {Corresponding scrambled Imagenette dataset with tile size 16}
    \label{fig_scrambled_imagenette}
\end{figure}

For instance, Figure \ref{fig_scrambled_imagenette} depicts the scrambling results of the imagenette dataset (Figure \ref{fig_imagenette}), where the tiles are 16x16 pixels.

A VGG16 model detailed in Section~\ref{cnn_architecture} was trained and tested on all six datasets, including both the original and scrambled versions. Table~\ref{table_scrambled_results} presents the classification accuracies for each model when trained on the scrambled datasets, compared to those trained on the original datasets.

\begin{table}[ht]
    \renewcommand{\arraystretch}{1.6} 
    \begin{center}
    \begin{tabular}[c]{|c|c|c|c|c|c|}
        \hline
        Dataset & Original accuracy & tile size = 1 & tile size = 16 & tile size = 32 & Random \\
        \hline
        Imagenette  & 59\%   & 10\%   & 28.4\% & 30.3\% & 10\%      \\
        Natural images & 85\% & 40.47\% & 36.3\% & 38.78\%  & 14.28\%       \\
        \hdashline
        CT scans    & 70\%   & 38\%   & 44\%   & 42\%   & 25\%      \\
        Coil-20     & 100\%  & 20\%   & 13\%   & 10\%   & 5\%       \\
        Yale Faces  & 70\%   & 18.8\% & 9\%    & 13.3\% & 6.66\%    \\
        \hdashline
        Caltech 20  & 36.6\% & 11.6\% & 14.1\% & 8.1\%  & 5\%       \\
        \hline
    \end{tabular}
    \caption{Classification accuracies of scrambled datasets}
    \label{table_scrambled_results}
   \end{center}
\end{table}

A consistent pattern is observed across all datasets and tile sizes: models trained on scrambled datasets still achieve much higher accuracy than random chance. Let us take a closer look at the results of a tile size of 1, when using a tile size of 1, where each pixel in an image is rearranged randomly, the image loses all recognizable structure and patterns, making it nearly impossible for humans to identify the class. Surprisingly, despite this, our VGG16 model manages to classify these scrambled images with unexpectedly high accuracy rates.

The Coil-20 dataset exhibits the highest bias or increase in accuracy compared to random guessing. Its random guessing accuracy of 5\% was improved by 400\%, reaching \(\sim\)20\%. The Yale Faces dataset shows the second highest increase, with accuracy rising from a random guessing rate of 6.6\% to \(\sim\)18.8\%, marking an increase of about 285\%. Other datasets also showed a high classification accuracy for scrambled images. In contrast, the Imagenette dataset displayed the least bias, with the scrambled accuracy matching the random guessing accuracy for a tile size of 1, which aligns with the expected behavior when images are scrambled.

\begin{figure}[ht]
    \includegraphics[width=6.5in]{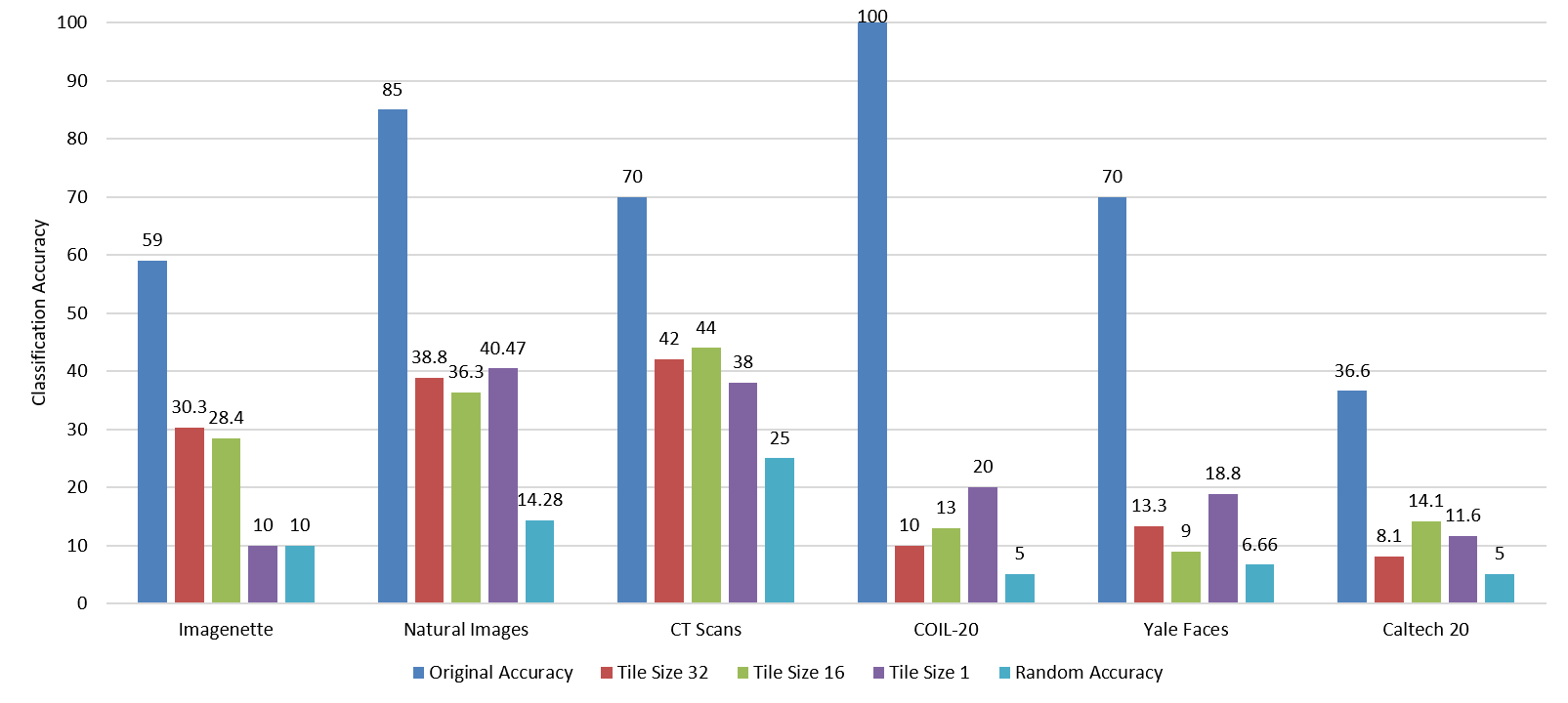}
    \caption[Classification accuracies of scrambled datasets vs random chance]
    {Classification accuracies of scrambled datasets vs random chance}
    \label{fig_scrambled_graph}
\end{figure}

In conclusion, our VGG16 models were able to accurately classify scrambled images, which would be impossible for humans to identify, as shown in Figure \ref{fig_scrambled_imagenette}. This indicates that nearly all datasets contain some level of bias, with the CNN memorizing irrelevant patterns and being influenced by these biases. Additionally, synthetic datasets such as Coil-20 and Yale Faces exhibited the highest level of bias compared to others.

\section{Conclusion}
\label{identifying_bias_conclusion}

We have analyzed the classification results for both methods: Cropping background segments (section \ref{cropping_background_segments}) and Image scrambling (section \ref{image_scrambling}). Both methods produced significantly higher accuracies compared to random chance. 

A 20$\times$20 pixel background segment, which typically lacks object features and is usually blank, should ideally not contribute to the neural network's learning and decision-making process. However, this is not the case here. The CNN achieved significantly higher accuracies than random chance (see Table \ref{table_cropped_results}), suggesting that these 20$\times$20 pixel cropped segments contain information not visible to the naked eye, which helps the model in distinguishing between image classes. Consequently, CNN appears to be learning from this hidden bias and relying on it for its final predictions.

When images are divided into tiles and scrambled randomly, predicting the object in the image becomes virtually impossible. Nonetheless, the model still achieves accuracy that is twice as high as random guessing, and in some cases, even higher (refer to Table \ref{table_scrambled_results}). This suggests that the model may be memorizing specific patterns and that a consistent bias is influencing its classification.

Results clearly show that these methods are highly effective in revealing dataset biases and any significant hidden background information that is influencing CNN's predictions. By applying these techniques, one can gain insight into the reliability of CNN classification accuracy and approach it with greater caution.

\cleardoublepage

\chapter{Identifying bias using image transforms}
\label{identifying_bias_image_transforms}

In Chapter \ref{identifying_bias_image_scrambling}, two techniques were discussed: cropped background segments and tile scrambling. These methods were effective in detecting general dataset biases. However, they struggled to differentiate between biases present in natural versus synthetic datasets.

Synthetic datasets are generated in a controlled environment, which theoretically reduces the likelihood of bias since all parameters are regulated. Despite this, biases often still occur. For instance, if all images of an object are captured in a single session, a strong and consistent bias can be introduced at the image acquisition stage itself. This capture bias is challenging to detect and can affect the CNN's classification accuracy. Consequently, the classification will be performed on session bias rather than focusing solely on the object of interest.

In contrast, natural datasets often face different challenges. Selection bias is particularly prevalent because images are typically collected from various sources. The methods used for filtering and selecting data can introduce selection bias, which in turn may result in representation bias by failing to fully capture the diversity or complexity of real-world scenarios.

In this chapter, we will aim to identify a method that effectively detects and differentiates between contextual information and the presence of background noise. We will apply a range of transformations, such as Fourier and Wavelet transforms, along with Median filtering, to the datasets. Following these pre-processing steps, we will train a CNN and evaluate how these transformations impact classification accuracy.

\section{Methods}
\label{reducing_methods}

\subsection{Fourier transform}
\label{fourier_transform}

The Fourier transform is one of the most widely employed methods in image processing to break down an image into its sine and cosine components. The transformation's outcome depicts the image in the frequency domain, whereas the input image corresponds to the spatial domain. In the resulting image, each point denotes a certain frequency present in the spatial domain image. The Fourier Transform is used in a wide range of applications, such as image analysis, image filtering, image reconstruction, and image compression.

As we are only concerned with digital images, we will restrict this discussion to the Discrete Fourier Transform (DFT). Since DFT is the sampled Fourier Transform, it does not contain all frequencies forming an image, but only a set of samples that is good enough to fully describe the image in the spatial domain. The number of frequencies equals the number of pixels in the spatial domain image, i.e. the image in the spatial and Fourier domains are of the same size \cite{fisher1996hypermedia}.

\begin{figure}[!ht]
    \centering
    \includegraphics[width=5.5in]{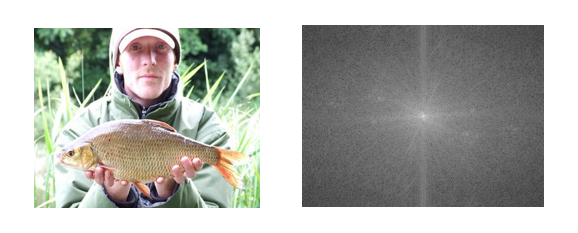}
    \caption[Fourier transformed image]
    {Fourier transformed image}
    \label{fig_fourier}
\end{figure}

\begin{figure}[!ht]
    \centering
    \includegraphics[width=3in]{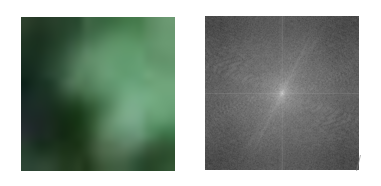}
    \caption[Fourier transform of 20$\times$20 cropped image]
    {Fourier transform of 20$\times$20 cropped image}
    \label{fig_fourier_cropped}
\end{figure}

As per OpenCV's documentation \cite{opencv_dft:2024}, the initial step involved converting all images to grayscale followed by applying the discrete Fourier transform. The resulting image in the Fourier domain is shown in Figure \ref{fig_fourier}. Along with the full images, the 20$\times$20 pixel cropped images were also transformed into Fourier domain, shown in Figure \ref{fig_fourier_cropped}.

\subsubsection{Results}
\label{fourier_results}

A VGG16 model has been trained and tested on the Fourier-transformed datasets and the classification accuracies are shown in Table \ref{table_fourier_results}.  

\begin{table}[ht]
    \setlength{\tabcolsep}{15pt}
    \renewcommand{\arraystretch}{1.6}
    \begin{center}
    \begin{tabular}{|*{6}{c|}}
        \hline
        Dataset & \multicolumn{2}{c|}{Full images} & \multicolumn{2}{c|}{Cropped images} & Random\\
        \hline
        & Original & Fourier & Original & Fourier & \\\cline{2-5}            
        Imagenette  & 59\%   & 38\%   & 17\%   & 10\% & 10\%      \\
        Natural images & 85\% & 76.8\%   & 27.6\% & 22\% & 14.28\%       \\
        \hdashline
        CT scans    & 70\%   & 54\%   & 44\%   & 40\% & 25\%      \\
        Coil-20     & 100\%  & 80\%   & 27.8\% & 26\% & 5\%       \\
        Yale Faces  & 70\%   & 60\%   & 17.7\% & 13\% & 6.66\%    \\
        \hdashline
        Caltech 20  & 36.6\% & 30\%   & 12.7\% & 12\% & 5\%       \\
        \hline
    \end{tabular}
    \caption{Classification accuracies of Fourier transformed datasets}
    \label{table_fourier_results}
   \end{center}
\end{table}

\begin{figure}[!ht]
    \centering
    \includegraphics[width=5in]{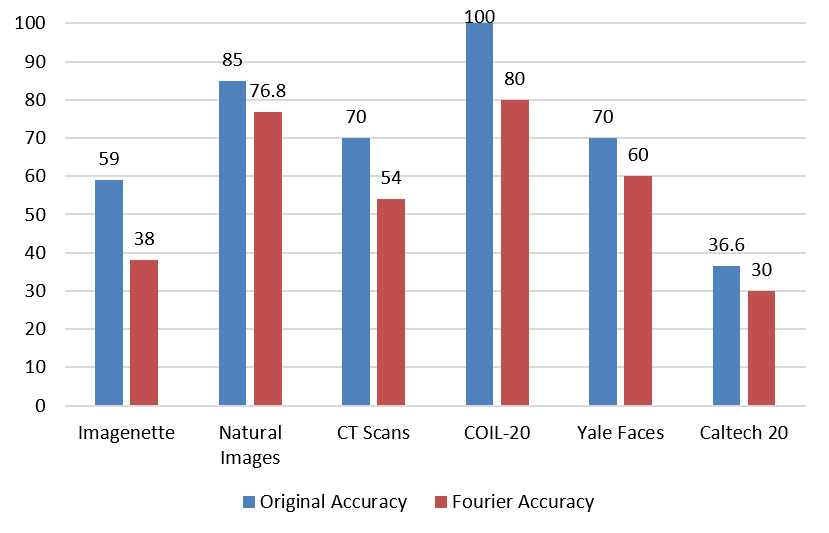}
    \caption[Classification accuracies of Fourier transformed full images]
    {Classification accuracies of Fourier transformed full images}
    \label{fig_ft_full_graph}
\end{figure}

\begin{figure}[!ht]
    \centering
    \includegraphics[width=5in]{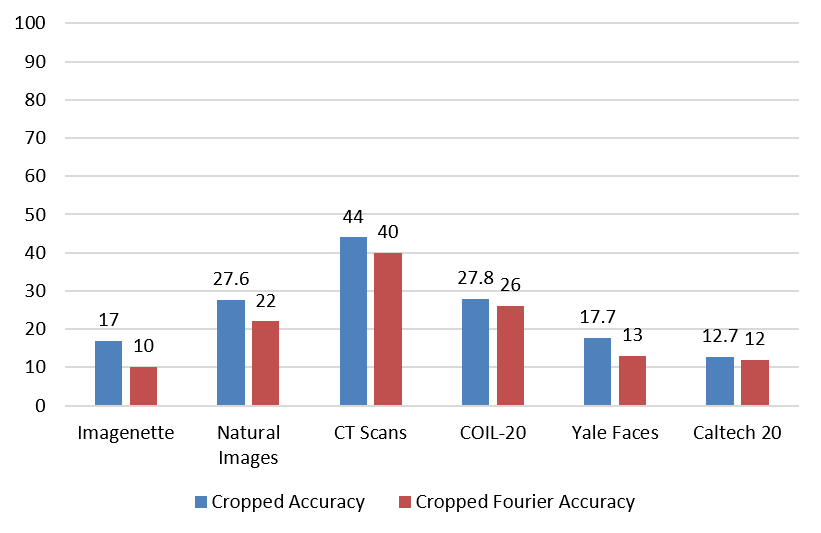}
    \caption[Classification accuracies of Fourier transformed cropped images]
    {Classification accuracies of Fourier transformed cropped images}
    \label{fig_ft_cropped_graph}
\end{figure}

A consistent trend is noticeable in the results: applying the Fourier transform has reduced the classification accuracy across all the datasets, both for full images (see Figure \ref{fig_ft_full_graph}) and 20$\times$20 cropped images (see Figure \ref{fig_ft_cropped_graph}). For example, in the Imagenette dataset, the accuracy dropped significantly from \(\sim\)59\% to \(\sim\)38\% for full images, and the accuracy for cropped images fell to random levels. Similarly, the Coil-20 dataset saw its accuracy decline from \(\sim\)100\% to \(\sim\)80\%. Not just these two datasets, but nearly all datasets experience a decrease in classification accuracy for both full and cropped images.

One possible explanation for this decrease could be that transforming images into the Fourier domain leads to a loss of spatial information. CNNs rely on this spatial information to effectively learn and classify features. It gets challenging to determine whether the drop in accuracy is due to the loss of relevant features or from the reduction of inherent bias in the datasets that were influencing CNN's predictions.

However, the Fourier transform does not effectively help in distinguishing between biases in natural and non-natural datasets, as it reduces accuracy across all types of datasets.


\subsection{Wavelet transform}
\label{wavelet_transform}

Wavelet transforms are mathematical methods used for investigating and retrieving data from images. \cite{agarwal2017analysis, othman2020applications} It breaks an image into components at various positions, preserving both frequency and spatial data. In contrast to the Fourier Transform, which outputs a global frequency, the Wavelet Transform provides a time-frequency (or space-frequency) representation, allowing for localized evaluation in both space and frequency domains. Wavelet transforms are presently being implemented to replace Fourier transforms in a wide range of applications, including medical imaging, image retrieval, watermarking, and many more.

\begin{figure}[!ht]
    \centering
    \includegraphics[width=6in]{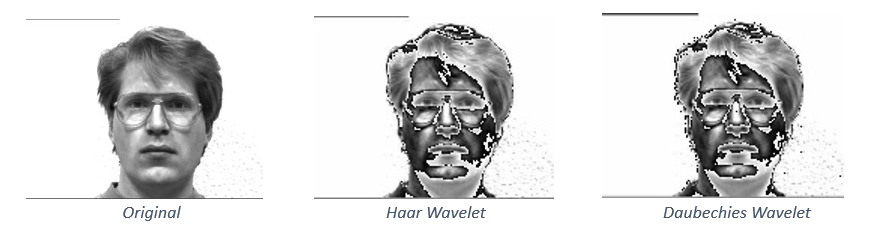}
    \caption[Discrete Wavelet transformed image]
    {Discrete Wavelet transformed image}
    \label{fig_wavelet}
\end{figure}

Discrete Wavelet Transform (DWT) is applied to all our datasets, full images along with the 20$\times$20 cropped ones, using a python package named PyWavelets \cite{pywt:2024}. Two different wavelets have been used in this study: Haar and Daubechies. Haar wavelets are discontinuous in nature and seem like a step function, while Daubechies wavelets best represent polynomial trends. Figure \ref{fig_wavelet} shows the images produced by applying these two wavelets, that appear quite comparable to one another.

\subsubsection{Results}
\label{wavelet_results}

A VGG16 model was trained and evaluated on datasets transformed using the Discrete Wavelet Transform, including both full and cropped images. Table \ref{table_haar_wavelet_results} shows the classification accuracies for models trained on Haar wavelet-transformed data. 

\begin{table}[ht]
    \setlength{\tabcolsep}{15pt}
    \renewcommand{\arraystretch}{1.6}
    \begin{center}
    \begin{tabular}{|*{6}{c|}}
        \hline
        Dataset & \multicolumn{2}{c|}{Full images} & \multicolumn{2}{c|}{Cropped images} & Random\\
        \hline
        & Original & Haar & Original & Haar & \\\cline{2-5}            
        Imagenette  & 59\%   & 50\%   & 17\%   & 13\% & 10\%      \\
        Natural images & 85\% & 72.77\%   & 27.6\% & 24.5\% & 14.28\%       \\
        \hdashline
        CT scans    & 70\%   & 69\%   & 44\%   & 56\% & 25\%      \\
        Coil-20     & 100\%  & 100\%  & 27.8\% & 31.4\% & 5\%       \\
        Yale Faces  & 70\%   & 80\%   & 17.7\% & 16\% & 6.66\%    \\
        \hdashline
        Caltech 20  & 36.6\%   & 35\%   & 12.7\% & 12.5\% & 5\%       \\
        \hline
    \end{tabular}
    \caption{Classification accuracies of Haar Wavelet transformed datasets}
    \label{table_haar_wavelet_results}
   \end{center}
\end{table}

\begin{figure}[!ht]
    \centering
    \includegraphics[width=5in]{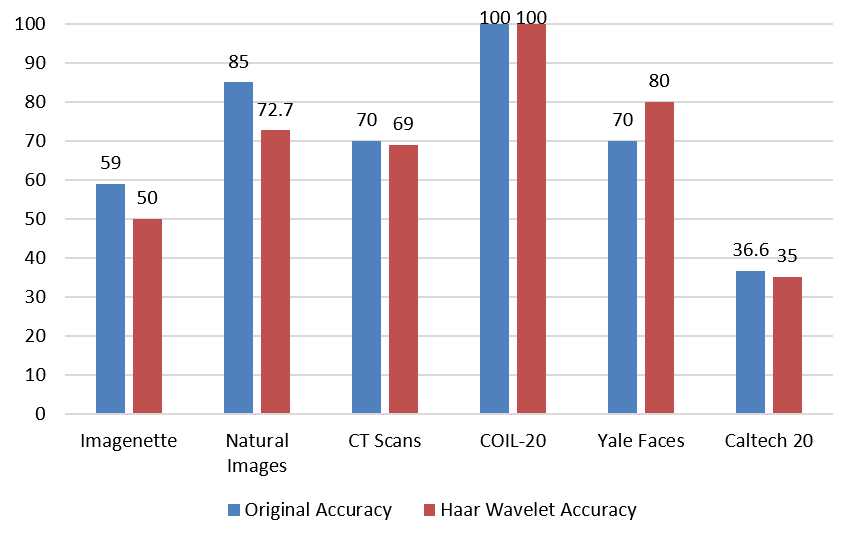}
    \caption[Classification accuracies of Wavelet transformed full images]
    {Classification accuracies of Wavelet transformed full images}
    \label{fig_wt_full_graph}
\end{figure}

The findings indicate a notable decline in prediction accuracies for natural datasets, while non-natural datasets have seen either consistent or improved accuracy rates. This trend is observed in both full images and cropped images. For instance, as shown in Figure \ref{fig_wt_full_graph} the prediction accuracy for the Imagenette dataset has dropped from \(\sim\)59\% to around \(\sim\)50\% for full images, and from \(\sim\)17\% to roughly \(\sim\)13\% (nearly random levels) for cropped images. In contrast, the Coil-20 dataset, which is synthetic, accuracy maintains a constant accuracy of 100\%, and its cropped image accuracy has increased from \(\sim\)27.8\% to about \(\sim\)31.4\%. 

The 20$\times$20 cropped images are usually blank and do not contain any visual information, but they still contain hidden information that can influence a CNN model's predictions. Applying a wavelet transform to these cropped images has increased their accuracy. This indicates that the Wavelet transform successfully revealed and enhanced the hidden signals in the background. Consequently, the bias in these non-natural datasets has risen. As a result, CNN has adapted to this increased bias and has become less reliable.

\begin{figure}[!ht]
    \centering
    \includegraphics[width=5in]{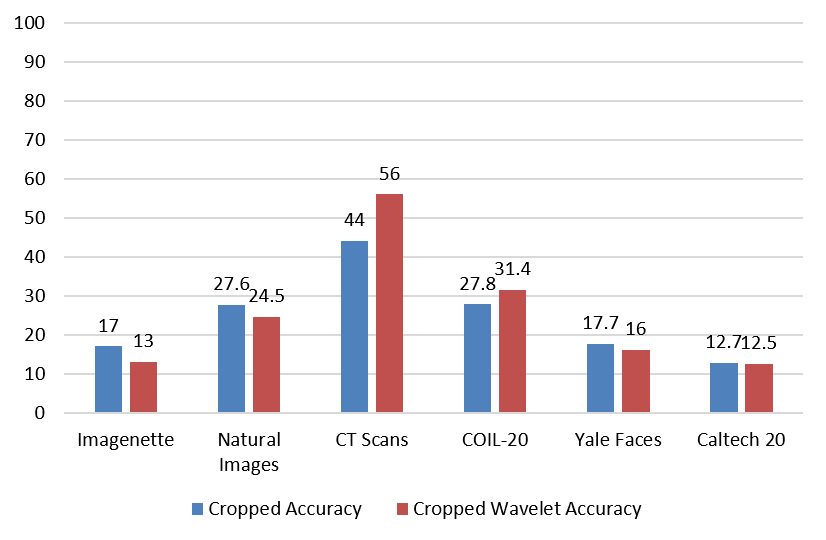}
    \caption[Classification accuracies of Wavelet transformed cropped images]
    {Classification accuracies of Wavelet transformed cropped images}
    \label{fig_wt_cropped_graph}
\end{figure}

In contrast, natural datasets typically do not contain capture bias and are therefore affected differently by the wavelet transform. As a result, the accuracies for Imagenette and Natural image datasets do not improve. We have now discovered a transform that effectively differentiates between contextual information and the presence of background noise and addresses them accordingly.

Table \ref{table_db_wavelet_results} presents the classification accuracies for models trained on data transformed using Daubechies wavelets. The performance of Daubechies wavelets is similar to that of Haar wavelets and also effectively differentiates between biases in natural and non-natural datasets.

\begin{table}[ht]
    \setlength{\tabcolsep}{11pt}
    \renewcommand{\arraystretch}{1.6}
    \begin{center}
    \begin{tabular}{|*{6}{c|}}
        \hline
        Dataset & \multicolumn{2}{c|}{Full images} & \multicolumn{2}{c|}{Cropped images} & Random\\
        \hline
        & Original & Daubechies & Original & Daubechies & \\\cline{2-5}            
        Imagenette  & 59\%   & 45.5\% & 17\%   & 13\% & 10\%      \\
        Natural images & 85\% & 72\%   & 27.6\% & 24\% & 14.28\%  \\
        \hdashline
        CT scans    & 70\%   & 69\%   & 44\%   & 56\% & 25\%      \\
        Coil-20     & 100\%  & 100\%  & 27.8\% & 30\% & 5\%       \\
        Yale Faces  & 70\%   & 80\%   & 17.7\% & 16\% & 6.66\%    \\
        \hdashline
        Caltech 20  & 36.6\% & 37.5\%   & 12.7\% & 12.7\% & 5\%     \\
        \hline
    \end{tabular}
    \caption{Classification accuracies of Daubechies Wavelet transformed datasets}
    \label{table_db_wavelet_results}
   \end{center}
\end{table}

The wavelet transform breaks down an image into numerous magnitudes, capturing features at various levels of detail. This breakdown highlights various frequencies such as approximation, and horizontal, vertical, and diagonal details, which can reveal edges, textures, and patterns. Compared to the Fourier transform, which only presents frequency details and not spatial data, the wavelet transform provides a time-frequency representation, allowing for localized examination in both the spatial and frequency domains. Additionally, the wavelet transform aids in noise reduction by adjusting wavelet coefficients to clean up images before they are fed into a classifier. These capabilities explain why the wavelet transform can effectively differentiate between contextual information and the presence of background noise and address them accordingly.

\subsection{Median Filter}
\label{median_filter}

A median filter is a non-linear image processing method that evaluates the image pixel by pixel and replaces each pixel with the median of adjacent entries. The pattern of nearby pixels depicts a window that slides across the entire image, one entry at a time. The window is usually a box or cross pattern, centered on the pixel being analyzed. \cite{DHANANI201319}

It is a smoothing technique that effectively removes disturbances from a noisy image or a signal. Unlike low-pass FIR filters, median filters are notable for retaining image edges, making them widely used in processing digital images.

\begin{figure}[!ht]
    \centering
    \includegraphics[width=5in]{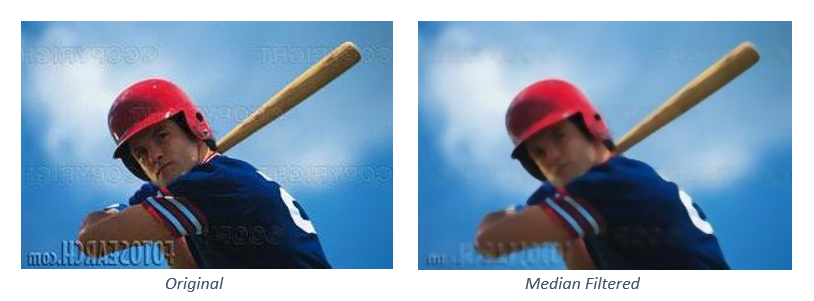}
    \caption[Median filtered image]
    {Median filtered image}
    \label{fig_median}
\end{figure}

Median filtering was performed on all our datasets using OpenCV's `medianBlur()` \cite{opencv_median:2024} method to smooth the images and reduce noise. Figure \ref{fig_median} shows the outcome of this process, highlighting the improved smoothness, reduced noise, and partial removal of watermarks.

\subsubsection{Results}
\label{median_results}

A CNN was trained and tested on datasets that had undergone median filtering, including both full and 20$\times$20 cropped background images. The classification accuracies are presented in Table \ref{table_median_results}. The results show that the Median filter can distinguish between contextual information and the presence of background noise. For full images, the accuracy of natural datasets has decreased; for example, Imagenette dropped from \(\sim\)59\% to about \(\sim\)55\%, and Natural images fell from \(\sim\)85\% to about \(\sim\)81.6\%. In contrast, the accuracy for synthetic datasets has remained stable, while hybrid datasets have improved. For instance, the accuracy for Caltech 20 increased from around  \(\sim\)36.6\% to around \(\sim\)39\%. 

\begin{table}[ht]
    \setlength{\tabcolsep}{15pt}
    \renewcommand{\arraystretch}{1.6}
    \begin{center}
    \begin{tabular}{|*{6}{c|}}
        \hline
        Dataset & \multicolumn{2}{c|}{Full images} & \multicolumn{2}{c|}{Cropped images} & Random\\
        \hline
        & Original & Median & Original & Median & \\\cline{2-5}            
        Imagenette  & 59\%   & 55\%   & 17\%   & 13.9\% & 10\%      \\
        Natural images & 85\% & 81.66\%   & 27.6\% & 23.5\% & 14.28\%       \\
        \hdashline
        CT scans    & 70\%   & 69\%   & 44\%   & 42\% & 25\%      \\
        Coil-20     & 100\%  & 100\%  & 27.8\% & 25\% & 5\%       \\
        Yale Faces  & 70\%   & 70\%   & 17.7\% & 17\% & 6.66\%    \\
        \hdashline
        Caltech 20  & 36.6\%   & 39\%   & 12.7\% & 12.2\% & 5\%       \\
        \hline
    \end{tabular}
    \caption{Classification accuracies of Median filtered datasets}
    \label{table_median_results}
   \end{center}
\end{table}

The median filter smooths the image and reduces noise, revealing that natural datasets had a visual bias that was eliminated by the filter. In contrast, synthetic datasets contain hidden information that the median filter cannot remove but can help reveal. As a result, the median filter addresses biases in natural and non-natural datasets differently.

\begin{figure}[!ht]
    \centering
    \includegraphics[width=5in]{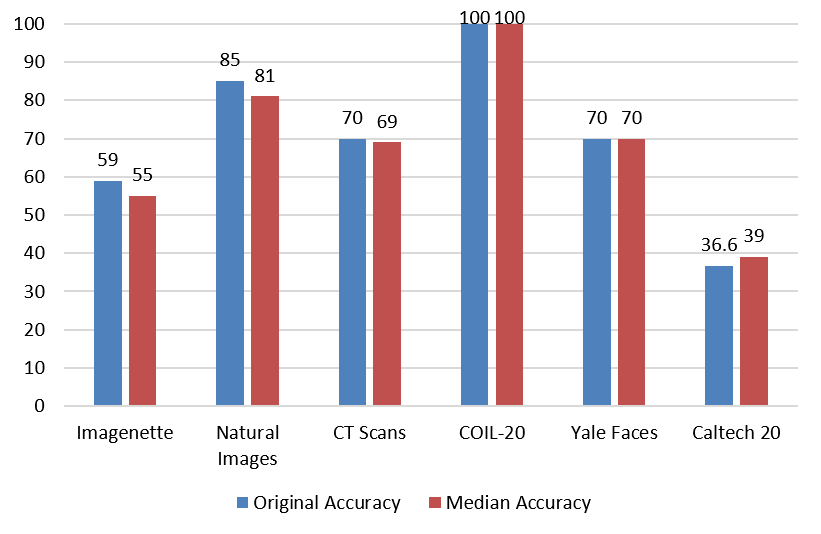}
    \caption[Classification accuracies of Median filtered full images]
    {Classification accuracies of Median filtered full images}
    \label{fig_median_full_graph}
\end{figure}

\begin{figure}[!ht]
    \centering
    \includegraphics[width=5in]{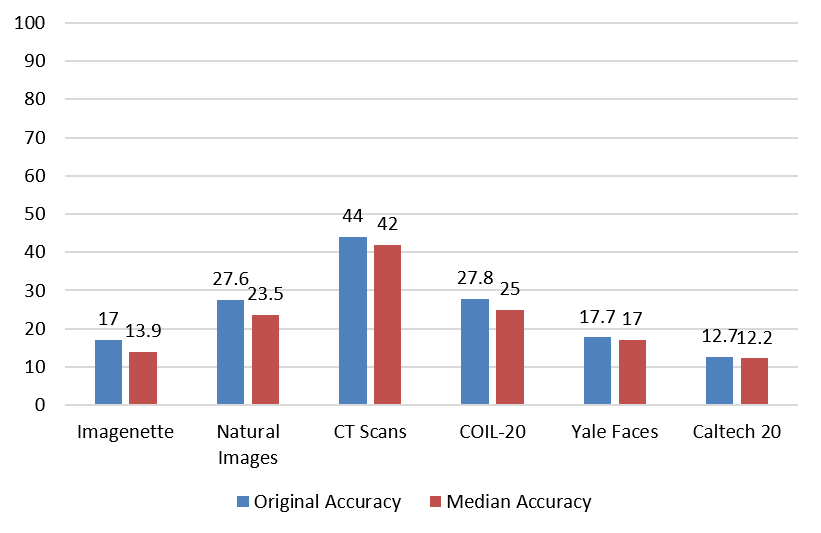}
    \caption[Classification accuracies of Median filtered cropped images]
    {Classification accuracies of Median filtered cropped images}
    \label{fig_median_cropped_graph}
\end{figure}

Since median filtering has proven to be effective and yielded promising results, we will now experiment by applying Fourier and Wavelet transforms on the median-filtered datasets. This will be covered in the subsequent sections. 

\subsection{Median filter + Fourier transform}
\label{median_fourier}

\begin{figure}[!ht]
    \centering
    \includegraphics[width=6in]{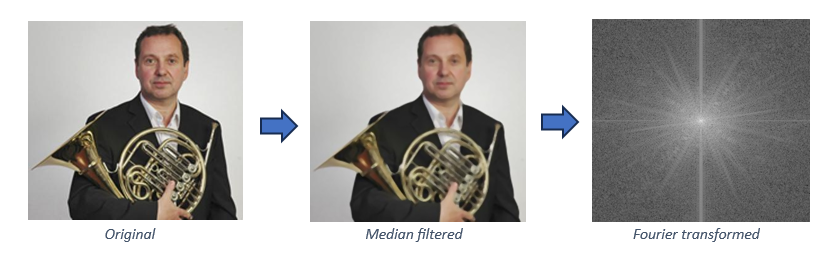}
    \caption[Median + Fourier transformed image]
    {Median + Fourier transformed image}
    \label{fig_median_fourier}
\end{figure}

Now, let's experiment by combining Median filtering and Fourier transform and examine see how it impact classification accuracies. First, median filtering is applied to the images to reduce noise and smooth them out. Next, the Fourier transform is applied to the median-filtered images and are converted into the Fourier domain.  Figure \ref{fig_median_fourier} displays the resultant image after applying the median filter and the Fourier transform.

Table \ref{table_median_fourier_results} shows the classification accuracies of VGG16 models trained and tested on the datasets, including both full images and 20$\times$20 cropped background images that have been median filtered and transformed into Fourier domain. We have observed a substantial and consistent decrease in accuracies across all datasets. For instance, Yale Faces and Imagenette are among those with the most significant reductions in accuracy.

\begin{table}[ht]
    \setlength{\tabcolsep}{8pt}
    \renewcommand{\arraystretch}{1.6}
    \begin{center}
    \begin{tabular}{|*{6}{c|}}
        \hline
        Dataset & \multicolumn{2}{c|}{Full images} & \multicolumn{2}{c|}{Cropped images} & Random\\
        \hline
        & Original & Median+Fourier & Original & Median+Fourier & \\\cline{2-5}            
        Imagenette  & 59\%   & 28\%   & 17\%   & 10\% & 10\%      \\
        Natural images & 85\% & 78.8\%   & 27.6\% & 16\% & 14.28\%       \\
        \hdashline
        CT scans    & 70\%   & 71.4\%   & 44\%   & 39.7\% & 25\%      \\
        Coil-20     & 100\%  & 58\%  & 27.8\%    & 26\% & 5\%       \\
        Yale Faces  & 70\%   & 6.6\%   & 17.7\%  & 13\% & 6.66\%    \\
        \hdashline
        Caltech 20  & 36.6\% & 30\%   & 12.7\%   & 14\% & 5\%       \\
        \hline
    \end{tabular}
    \caption{Classification accuracies of Median filtered Fourier transformed datasets}
    \label{table_median_fourier_results}
   \end{center}
\end{table}

Earlier, we observed that converting images to the Fourier domain results in some loss of spatial information, leading to a decrease in accuracy. Additionally, the median filter smooths the images and reduces noise, which also contributes to a modest decline in accuracy. Combining these two techniques is expected to cause a more significant drop in accuracy.

\begin{figure}[!ht]
    \centering
    \includegraphics[width=5in]{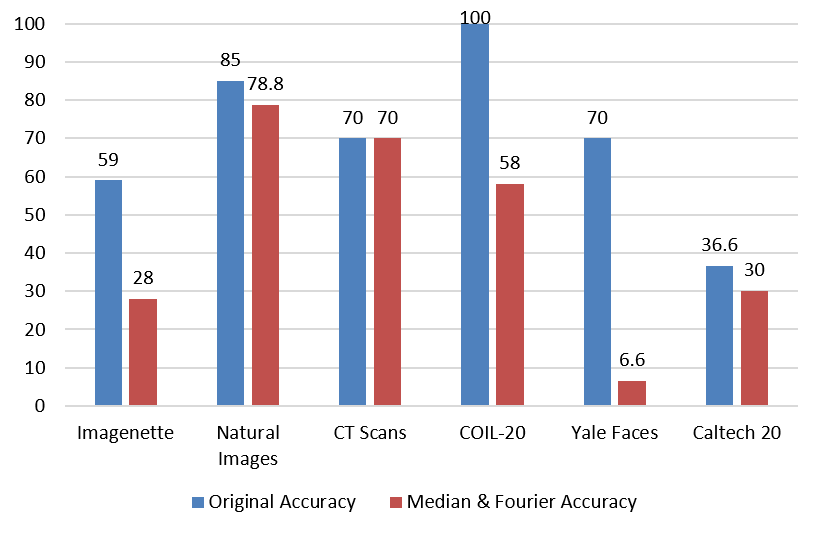}
    \caption[Classification accuracies of Median and Fourier transformed full images]
    {Classification accuracies of Median and Fourier transformed full images}
    \label{fig_mft_full_graph}
\end{figure}

\begin{figure}[!ht]
    \centering
    \includegraphics[width=5in]{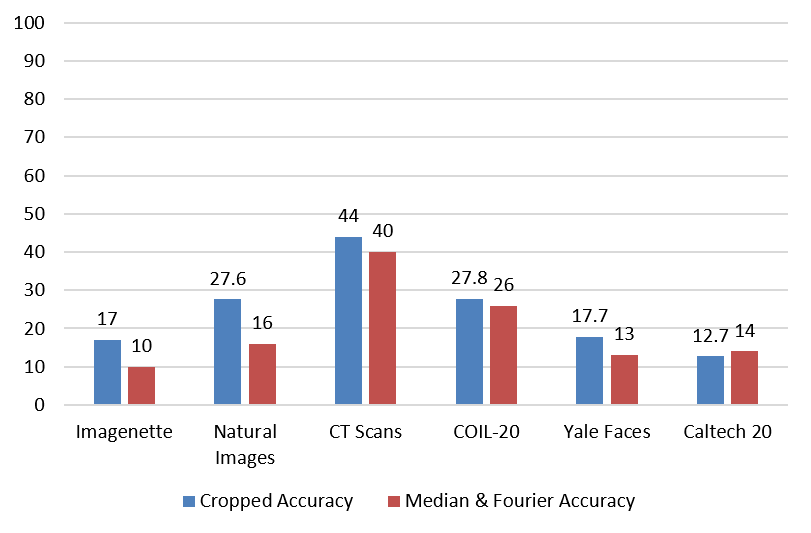}
    \caption[Classification accuracies of Median and Fourier transformed cropped images]
    {Classification accuracies of Median and Fourier transformed cropped images}
    \label{fig_mft_cropped_graph}
\end{figure}

Since the Fourier transform was not particularly effective at distinguishing between biases in natural and synthetic datasets, combining it with the median filter also proved ineffective. This combination did not differentiate but instead led to a decrease in accuracies across nearly all datasets.



\subsection{Median filter + Wavelet transform}
\label{median_wavelet}

\begin{figure}[!ht]
    \centering
    \includegraphics[width=6in]{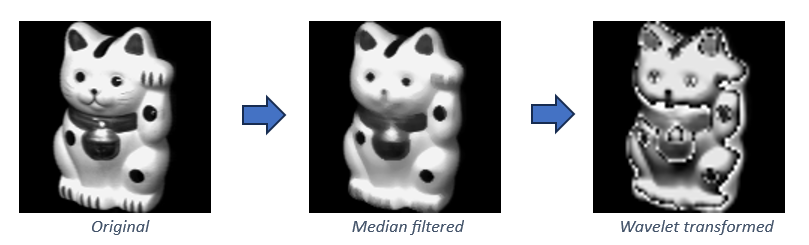}
    \caption[Median + Wavelet transformed image]
    {Median + Wavelet transformed image}
    \label{fig_median_wavelet}
\end{figure}

In this section, we will investigate the effects of combining median filtering with the Haar wavelet transform. First, median filtering was applied to all datasets to smooth the images and reduce noise. Following this, the images were converted into the space-frequency domain using the Haar wavelet transform. The resulting images, after both median filtering and the Haar wavelet transform, are displayed in Figure \ref{fig_median_wavelet}. This approach aims to leverage the benefits of both methods.

Table \ref{table_median_wavelet_results} presents the classification accuracies for VGG16 models trained and tested on datasets pre-processed with both median filtering and the Haar wavelet transform. We previously observed that median filtering and wavelet transforms were each effective in identifying and distinguishing contextual information and the presence of background noise. Combining these two methods has also proven to be highly successful.

\begin{table}[ht]
    \setlength{\tabcolsep}{8pt}
    \renewcommand{\arraystretch}{1.6}
    \begin{center}
    \begin{tabular}{|*{6}{c|}}
        \hline
        Dataset & \multicolumn{2}{c|}{Full images} & \multicolumn{2}{c|}{Cropped images} & Random\\
        \hline
        & Original & Median+Wavelet & Original & Median+Wavelet & \\\cline{2-5}            
        Imagenette  & 59\%   & 49.5\%   & 17\%   & 14\% & 10\%      \\
        Natural images & 85\% & 70.2\%   & 27.6\% & 22.5\% & 14.28\%       \\
        \hdashline
        CT scans    & 70\%   & 70.5\%   & 44\%   & 33.4\% & 25\%      \\
        Coil-20     & 100\%  & 100\%  & 27.8\%    & 30\% & 5\%       \\
        Yale Faces  & 70\%   & 80\%   & 17.7\%  & 11\% & 6.66\%    \\
        \hdashline
        Caltech 20  & 36.6\% & 36\%   & 12.7\%   & 13\% & 5\%       \\
        \hline
    \end{tabular}
    \caption{Classification accuracies of Median filtered Wavelet transformed datasets}
    \label{table_median_wavelet_results}
   \end{center}
\end{table}

The accuracies for natural datasets, such as Imagenette and Natural images, decreased by about \(\sim\)9.5\% and \(\sim\)14.8\% respectively. While those for synthetic and hybrid datasets either remained stable or increased. This can be attributed to the median filter, which smooths and reduces noise in the images, and the wavelet transform, which identifies and enhances the capture bias (hidden information present in the background introduced during the image acquisition stage) in synthetic datasets. This pairing accounts for the observed changes in classification accuracies.

\begin{figure}[!ht]
    \centering
    \includegraphics[width=4.7in]{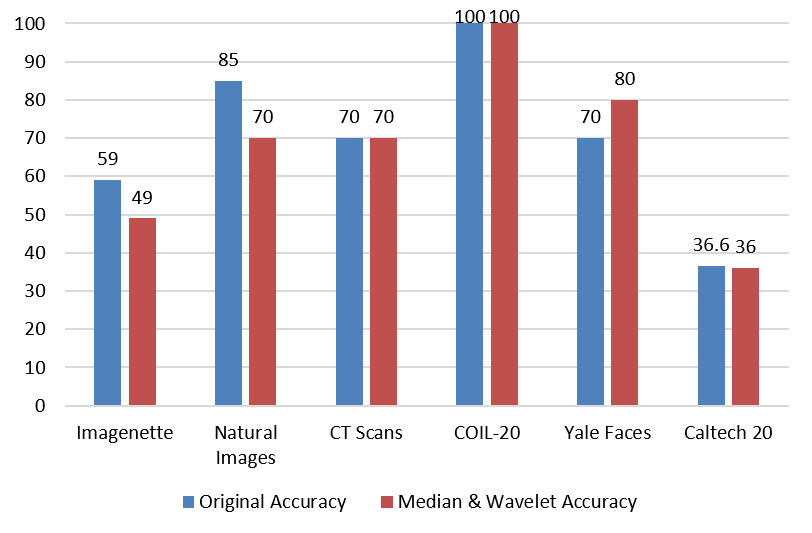}
    \caption[Classification accuracies of Median and Wavelet transformed full images]
    {Classification accuracies of Median and Wavelet transformed full images}
    \label{fig_mwt_full_graph}
\end{figure}

\begin{figure}[!ht]
    \centering
    \includegraphics[width=4.7in]{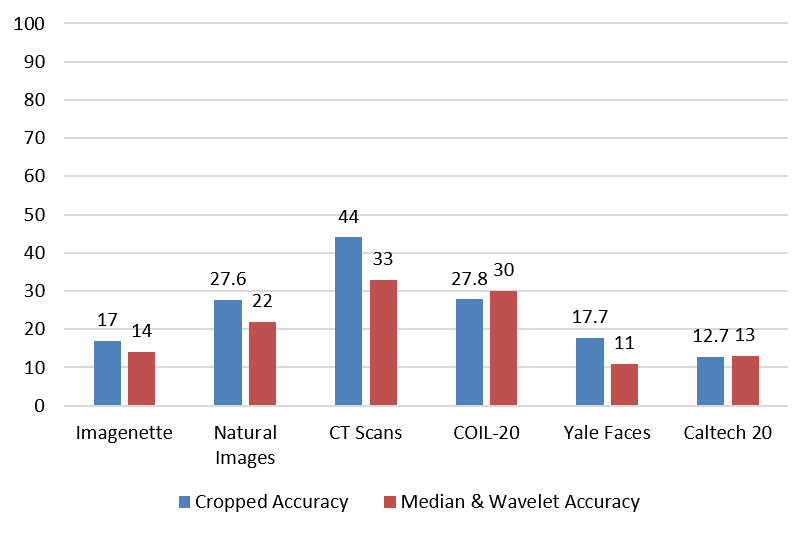}
    \caption[Classification accuracies of Median and Wavelet transformed cropped images]
    {Classification accuracies of Median and Wavelet transformed cropped images}
    \label{fig_mwt_cropped_graph}
\end{figure}

\section{Conclusion}
\label{reducing_bias_conclusion}

In Section \ref{reducing_methods}, we reviewed several techniques, including Fourier, Wavelet transforms, Median filtering, and their combinations, which were used to identify bias and distinguish between contextual information and the presence of background noise. The Wavelet transform \ref{wavelet_transform}, Median filtering \ref{median_filter}, and their combination \ref{median_wavelet} have all proven highly effective in distinguishing between biases in natural and non-natural datasets. The results show that the method can distinguish between contextual information and the presence of background noise, and alert on the presence of background noise even without the need to use background information.

Applying a median filter smooths the image and eliminates noise, and subsequently applying Wavelet transform to this smoothed dataset further enhances the signal in the image. Notably, the wavelet transform also highlights any hidden signals introduced during the image acquisition process. In a nutshell, these methods decreased the prediction accuracies for natural datasets but improved the accuracies for non-natural datasets, effectively addressing the distinct biases present in each.

Before fully trusting the high prediction accuracies of CNNs, it is strongly recommended to apply the methods outlined in Chapter \ref{identifying_bias_image_scrambling} to determine if there is any persistent bias present in the dataset. Then the techniques discussed in Chapter \ref{identifying_bias_image_transforms} should help you distinguish between contextual information and the presence of background noise. This evaluation helps gauge the reliability of the CNN and ensures that its accuracy is trustworthy.

\newpage
\section{Code repository}
\label{code_repository}

The complete code utilized in this research study is available in a GitHub repository:
\href{https://github.com/SaiTeja-Erukude/identifying-bias-in-dnn-classification}{https://github.com/SaiTeja-Erukude/identifying-bias-in-dnn-classification}. \cite{erukude:2024}

\subsection{Files description}
\label{files_description}

vgg16\_model.py: This file contains the architecture of the VGG16 neural network, which was used for model training.\\

Within the preprocess folder, the following files are included:

\begin{itemize}
    \item augment.py: This script is responsible for augmenting images from the Yale Faces dataset.
    \item crop.py: This file implements the cropping background segments method discussed in Section \ref{cropping_background_segments}.
    \item scramble.py: This script executes the ``Tile Scrambling" technique, which randomly shuffles the pixels of the images as outlined in Section \ref{image_scrambling}.
    \item fourier.py: This script applies the Fourier transform to all images located in a specified directory.
    \item wavelet.py: This script applies wavelet transformations to images. It can accommodate various types of wavelets by simply modifying the type specified on line 46.
    \item median.py: This script uses OpenCV’s medianBlur() method to apply median filtering with a window size of 5.
\end{itemize}

Within the predict folder, the following files are included:
\begin{itemize}
    \item predict.py: This script is used for importing the trained model and evaluating its performance on a set of test images. 
    \item ensemble\_predict.py: This script creates an ensemble classifier that incorporates models trained on the original images, as well as those transformed by Fourier and wavelet techniques.
\end{itemize}





\cleardoublepage
\phantomsection


\addcontentsline{toc}{chapter}{Bibliography}
\bibdata{references}
\bibliography{references}



\end{document}